\documentclass[letter]{article}

\usepackage{lipsum}
\usepackage{amsfonts}
\usepackage{graphicx}
\usepackage{epstopdf}
\usepackage[caption=false]{subfig}
\usepackage{wrapfig}
\usepackage{color}
\usepackage{amsmath}
\usepackage[compact]{titlesec}
\usepackage{enumitem}
\setlist[enumerate]{leftmargin=.28in}
\setlist[itemize]{leftmargin=.28in}
\usepackage[font=small,textfont=it,labelfont=bf]{caption}

\usepackage{algorithm}
\usepackage[noend]{algpseudocode}

\DeclareMathOperator*{\argmin}{arg\,min}

\usepackage{mathtools}
\usepackage{calc}
\newcommand*{\mybar}[2][0pt]{%
  \setbox0=\hbox{$#2$}%
  \overline{\mathrlap{\phantom{\rule{\wd0}{\ht0+{#1}}}}\smash{#2}}%
}
\newcommand{\avld}{\smash{\mybar[-0.25pt]{\mathbbm{d}}}}

\newcommand{\lD}{\mathbbm{D}}
\newcommand{\ld}{\mathbf{d}}

\newcommand{\changes}{}

\usepackage{authblk}

\title{Parallel Transport Unfolding:\\ \large A Connection-based Manifold Learning Approach}

\author[1]{\small Max Budninskiy}
\affil[1]{\footnotesize Caltech}
\affil[2]{Michigan State University}
\author[1]{Gloria Yin}
\author[1]{Leman Feng}
\author[2]{Yiying Tong}
\author[1]{Mathieu Desbrun}

\date{\vspace*{-5mm}\footnotesize June 21 2018\vspace*{-6mm}}

\usepackage{amsmath,amssymb,txfonts,bbm}
\DeclareSymbolFont{bbold}{U}{bbold}{m}{n}
\DeclareSymbolFontAlphabet{\mathbbold}{bbold}
\usepackage{amsopn}

\usepackage{hyperref}
\hypersetup{
  pdftitle={Parallel Transport Unfolding: A Connection-based Manifold Learning Approach},
  pdfauthor={M. Budninskiy, G. Yin, L. Feng, Y. Tong, and M. Desbrun},
	colorlinks=true,linkcolor=magenta,citecolor=blue,urlcolor=blue,breaklinks=true
}
\usepackage[hyphenbreaks]{breakurl}
\usepackage{breakcites}

\begin{document}

\maketitle

\begin{abstract}
Manifold learning offers nonlinear dimensionality reduction of high-dimensional datasets. In this paper, we bring geometry processing to bear on manifold learning by introducing a new approach based on metric connection for generating a quasi-isometric, low-dimensional mapping from a sparse and irregular sampling of an arbitrary manifold embedded in a high-dimensional space. Geodesic distances of discrete paths \changes{over} the input pointset are evaluated through ``parallel transport unfolding'' (PTU) to offer robustness to poor sampling and arbitrary topology.
Our new geometric procedure exhibits the same strong resilience to noise as one of the staples of manifold learning, the Isomap algorithm, as it also exploits all pairwise geodesic distances to compute a low-dimensional embedding.
While Isomap is limited to geodesically-convex sampled domains, parallel transport unfolding does not suffer from this crippling limitation, resulting in an improved robustness to irregularity and voids in the sampling.
Moreover, it involves only simple linear algebra, significantly improves the accuracy of all pairwise geodesic distance approximations, and has the same computational complexity as Isomap. Finally, we show that our connection-based distance estimation can be used for faster variants of Isomap such as L-Isomap.
\end{abstract}



\section{Introduction}

With the avalanche of information in our big data era, the demand for numerical tools to analyze and classify high-dimensional data has become paramount. Recognizing and exploiting hidden structures present in the input is particularly desirable when reducing the dimensionality of the data~\cite{Vidal:2016:GPC}. The manifold assumption, which posits that many high-dimensional datasets actually live on much lower dimensional curved spaces, has been proven surprisingly useful in a variety of contexts. Algorithms working towards extracting reduced dimensional representations of data, also called \emph{manifold learning}, produce a low count of ``intrinsic variables'' to describe the high-dimensional input---with which markedly faster computations can be performed in the  context of machine learning (to address the ``curse of dimensionality''),  facial animation~\cite{Kim2007}, skeletal animation~\cite{Assa:2005:ASP}, video editing~\cite{Pless:2003}, and nonlinear shape editing~\cite{Max:2017} to mention a few.\\[-4mm]

Perhaps the most popular algorithm for dimensionality reduction is Principal Component Analysis (PCA); yet, it is only useful when data lie on or close to a linear subspace of the high-dimensional space. Manifold learning algorithms can be viewed as nonlinear extensions to PCA:
since data points are assumed to be samples from a low-dimensional manifold that is embedded in a high-dimensional space, many nonlinear dimensionality reduction algorithms attempt to uncover a quasi-isometric parametrization of this manifold. More precisely,  they seek an as-isometric-as-possible mapping from the input pointset in $\varmathbb{R}^D$ into a low-dimensional space $\varmathbb{R}^d$ (with $d\!\ll\!D$) that represents a low-distortion unfolding of the original data. 
\changes{This is, therefore, the high-dimensional counterpart of \emph{parameterization} in computer graphics, a task  that finds a $(u,v)$ parameterization for a two-dimensional patch embedded in 3D (e.g., to texture its surface) and for which a variety of approaches exploiting the three-dimensional nature of the embedding space have been proposed~\cite{Hormann:2008:MPT,ParamMD:2003,Mullen:2008:SCP,Param:2009}.} 
\smallskip

In this paper, we bring discrete differential geometry concepts to bear on this ubiquitous and inherently geometric issue, and show how current limitations on geodesic convexity of the input manifold can be lifted elegantly using the Levi-Civita connection, for the same computational complexity.

\begin{figure}[t]
  \centering
  \includegraphics[width=0.9\linewidth]{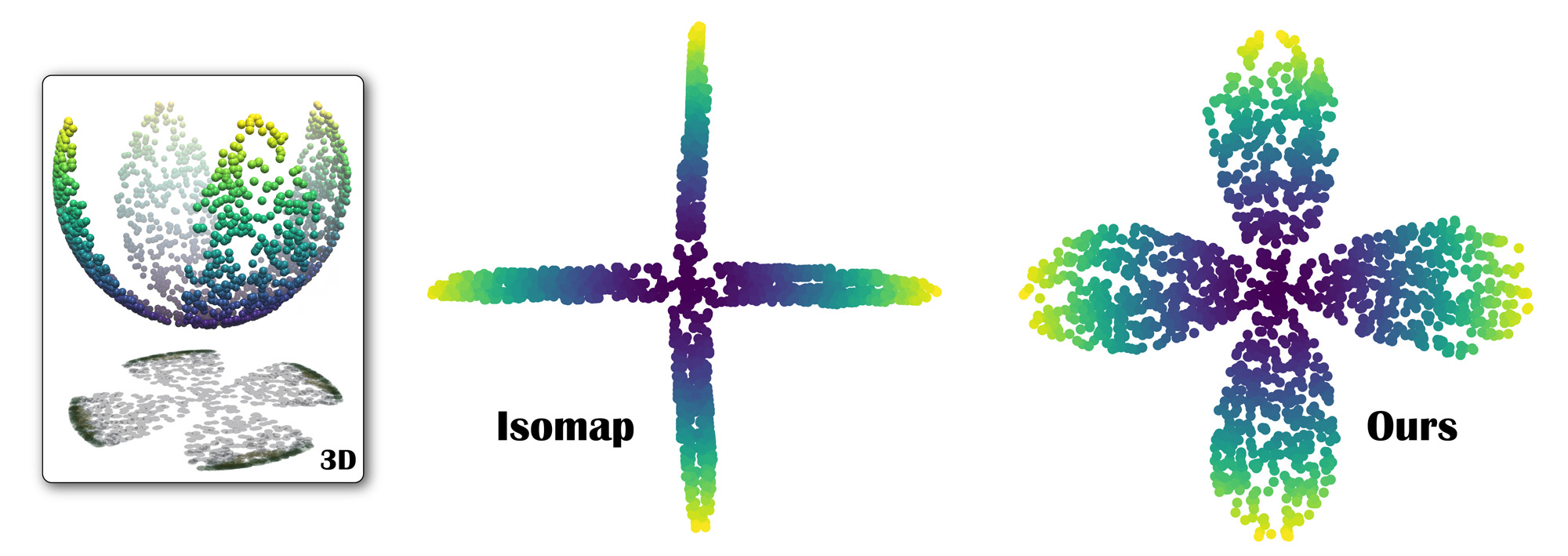}\vspace*{-4mm}
	\caption{\textbf{Correctly Unfolding Petals.} For a 3D sampling of a 4-petal shaped portion of a sphere (left), Isomap (a staple of manifold learning) fails to find a near isometric 2D parameterization (middle) due to \changes{geodesic} non-convexity of the intrinsic geometry. Our parallel transport approach, instead, handles this case as expected (right). Lifting the pointset to much higher dimensions and applying random rotations and reflections would not change these results. \vspace*{-2mm}}
	\label{fig:Petals}
\end{figure}

\subsection{Previous Work}
There are two distinct flavors of manifold learning in existing work: local methods, which only rely on local measurements and sparse matrix eigenanalysis to recover the low-dimensional manifold, and global methods, that leverage intrinsic measurements between all data points and involve eigenanalysis of a dense matrix.
\vspace*{1mm}

\paragraph*{Local methods.}
Local embedding methods (e.g., Laplacian Eigenmaps~\cite{LaplacianEig}, Locally Linear Embeddings~\cite{Roweis:2000:LLE} and their many variants) consider local neighborhoods of input points to infer the whole structure of the data through global stitching of these neighborhoods: \changes{for each of the $n$ input points, they determine its relative position with respect to its immediate surroundings}, and then rely on a sparse matrix eigenanalysis to deduce a global low-dimensional embedding that best preserves local relative positions. While no restrictive assumptions on the geometry or topology of the manifold are required, this approach to nonlinear dimensionality reduction often generates large distortion in the resulting embedding---an issue fortunately addressed by enforcing
local linear precision, see~\cite{Donoho:2003,Zhang:2006:MLLE,Max:2017}. The local nature of this family of methods typically implies a computational complexity of $\mathcal{O}(n^{1.5})$ due to the sparsity of the matrix involved in the eigenanalysis. Yet this same locality hampers its robustness: the presence of noise and outliers often leads to degenerate results~\cite{Maaten:2009:DR}.
Local embedding methods are thus general and efficient enough to handle arbitrary inputs, but are often not resilient enough to noise to be practical (see Figs.~\ref{fig:localFail} and~\ref{fig:Petals2} for instance).
\vspace*{1mm}

\paragraph*{Global methods.}
The Isomap technique~\cite{isomap} is a variant of MultiDimensional Scaling (MDS)~\cite{Torgerson:1965, Crippen:1978} that attempts to reduce distortion in the mapping by preserving pairwise \emph{geodesic} distances between all data points as well as possible. After forming a $k$-nearest neighbor graph of the data points as a representation of the input manifold, Isomap first solves the all-pairs shortest path problem (using Dijkstra's or Floyd-Warshall algorithm) to approximate geodesic distances between every pair of points in the graph. Finally, a low-dimensional embedding that best preserves these pairwise distances is then constructed with MDS through eigenanalysis of the corresponding Gram matrix~\cite{MDSBook}. The high computational complexity due to the eigenanalysis of a dense Gram matrix (in $\mathcal{O}(n^{3})$) can be further improved through probabilistic linear algebra~\cite{Halko:2011}, variants such as L-Isomap~\cite{l_isomap} or by exploiting various numerical improvements of MDS~\cite{Harel:2002,Brandes:2007,Khoury:2012} that approximate its solution in near linear time. Since \textit{all} pairwise geodesic distances are used, Isomap is particularly robust to noise, and for this reason it remains one of the most popular algorithms used for manifold learning.

\subsection{Shortcomings of Isomap}
\begin{figure}[t]
\begin{minipage}{0.49\linewidth}
  \centering
	\includegraphics[width=\linewidth]{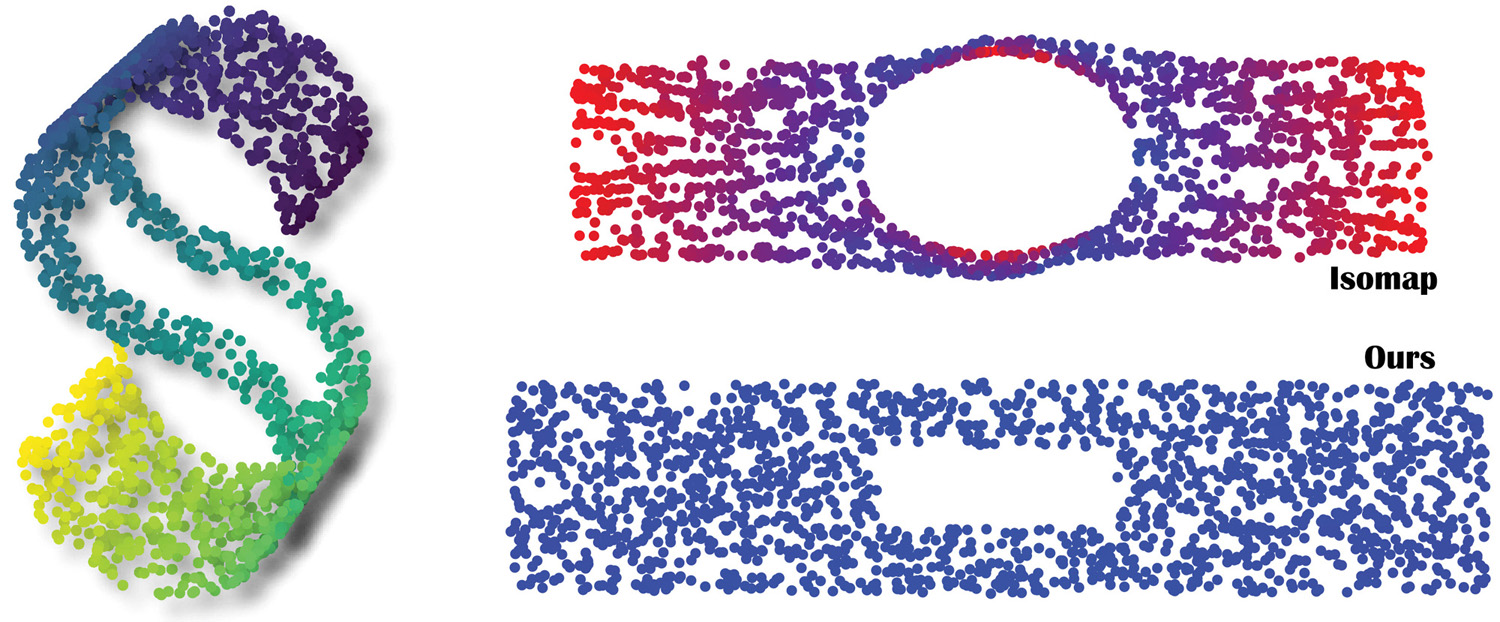}\vspace*{-3mm}
	\caption{\textbf{Holey S.}
		Isomap (top right) fails to find a quasi-isometric embedding of a uniformly sampled developable S-shaped $2$-manifold with a sampling void in the middle (left) in 3D. Our approach (bottom right) unfolds it almost perfectly. A color ramp (blue: 0\% error, to red: 15\% error) is used in the visualization of the two embeddings.}
	\label{fig:holyS}
\end{minipage}
\hfill
\begin{minipage}{0.49\linewidth}
\centering
	\includegraphics[width=\columnwidth]{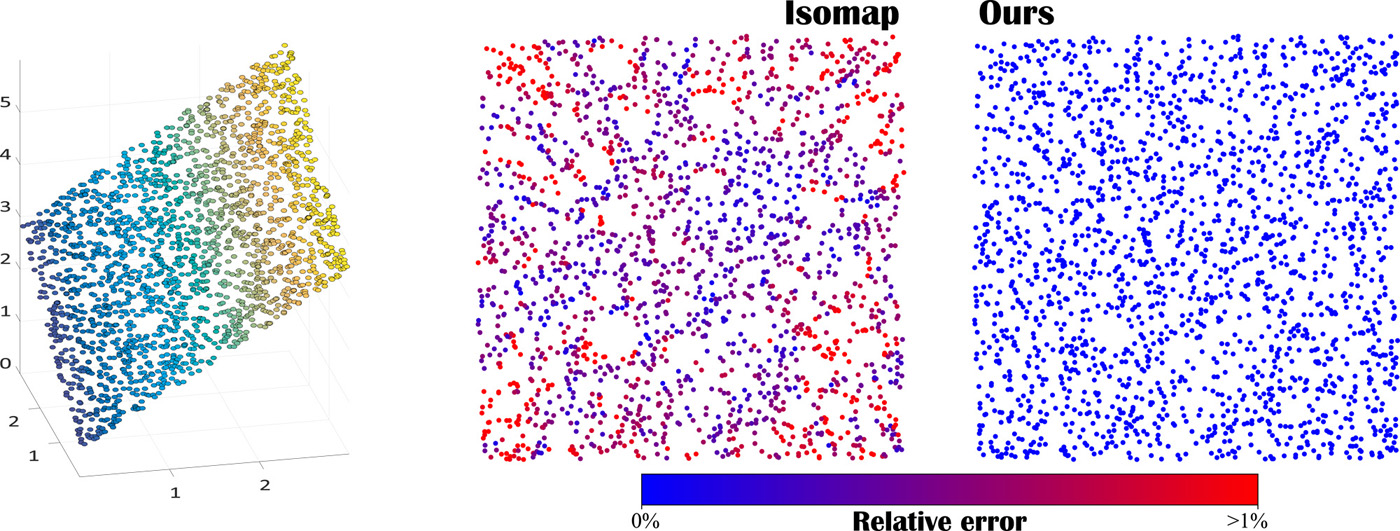}\vspace*{-2.5mm}
	\caption{\textbf{Linear Precision.} Due to its reliance on graph-based paths, Isomap distorts a flat 2D pointset embedded in 3D; instead, our connection-based approach (right) flattens it exactly. A linear color ramp from blue ($0\%$ $\ell^2$-error in position relative to bbox size) to red ($>\!1\%$ error) is used in the visualization of the two embeddings.}
	\label{fig:notAffinePrecise}
\end{minipage}
\vspace*{-4mm}
\end{figure}

Despite its resilience to noise, Isomap lacks severely in generality: it can \emph{only} handle geodesically convex regions correctly. In particular, if the input sampling of the manifold has a large void, Isomap will significantly overestimate geodesic distances since graph-based shortest paths for points on opposite side of the hole will be forced to go around the empty region, thus resulting in a distorted embedding (Fig.~\ref{fig:holyS}). This issue is all the more serious as it impacts irregular sampling in general: unlike PCA, Isomap fails to produce a non-distorted embedding for a dataset lying precisely on a linear subspace of the ambient high-dimensional space, since graph-based shortest paths are only approximations of true manifold geodesics, see Fig.~\ref{fig:notAffinePrecise}. There have been attempts to alleviate this limitation, such as the Topologically Constrained Isometric Embedding (TCIE) approach~\cite{Rosman:2010:TCIE} which first tries to detect the boundaries of holes and the geodesic paths that go through these boundaries, before constructing a map that ignores these biased geodesic paths. However, the process of detecting boundaries is highly susceptible to noise, since it can often be unclear whether an empty region delimits a ``real'' void or whether the local sampling is merely low. Moreover, the subsequent minimization is significantly more difficult to perform as it no longer relies on a simple partial eigendecomposition. Consequently Isomap is often used in practice, even when the topology and geometry of the sampled manifold are not known---because of the lack of another robust learning approach to use instead.

\subsection{Overview and contributions}
In this paper we introduce Parallel Transport Unfolding (PTU), a manifold learning technique that performs nonlinear dimensionality reduction to produce a quasi-isometric, low-dimensional embedding of an arbitrary set of high-dimensional data points. By simply replacing the Dijkstra path-based geodesic distance estimates with parallel transport based approximations instead, our approach removes the geometric limitations of Isomap: it can reliably handle arbitrary data with strongly irregular sampling while retaining its resilience to noise. We show that this new geometric procedure no longer requires geodesic convexity of the domain sampled by the input data, only involves simple linear algebra, significantly improves the accuracy of all pairwise geodesic distance approximations, and does not change the overall computational complexity of the original Isomap procedure. \changes{Moreover, our approach exploits the low dimensionality of the manifold by using a connection on the $d$-dimensional tangent bundle for efficiency purposes: this markedly differs from previous parallel transport based geodesic computations on 2-manifolds in 3D, where the codimension is $1$ and thus, where either the normal field or 2D polar coordinates can be leveraged to derive fast geodesic approximations~\cite{Schmidt:2006,Melvaer:2012,Schmidt:2013}.}
Finally, we also demonstrate that our connection-based distance estimation applies equally well to Landmark-Isomap~\cite{l_isomap}, a variant offering significant improvements in computational time but suffering from the same convexity limitation as Isomap.

\subsection{Notations and assumptions}
Throughout our exposition, we consider an input data set $\mathcal{S}\!=\!\{\mathbf{x}_i \}_{i=1..n}$ of $n$ points that irregularly sample (possibly with noise) a connected compact $d$-manifold $\mathcal{M}$ embedded in a $D$-dimensional metric space. While the use of arbitrary kernels has been proposed as a unifying view~\cite{Ham:2004:KVD,Weinberger:2004:LKM}, we will assume the embedding space to be $\varmathbb{R}^D$ for simplicity, with $d \!\ll\!D$. We only assume that $\mathcal{M}$ possesses an atlas with a single chart, so that an injective $d$-dimensional parametrization of the manifold exists. Our parallel transport based manifold learning algorithm seeks to find an as-isometric-as-possible $d$-dimensional embedding of the input $\mathcal{S}$ as a pointset $\{\mathbf{z}_i \}_{i=1..n} \!\!\subset\!\! \mathbf{R}^d$. Finally, to simplify further expressions, we assemble a $D \times n$ matrix $\mathbf{X} \!=\! \left(\mathbf{x}_1, ... ,\mathbf{x}_n\right)$ from the input points and denote the final embedding as a $d \times n$ matrix $\mathbf{Z} \!=\! (\mathbf{z}_1, ... , \mathbf{z}_n)$.

\section{Primer on Isomap}
Before describing our method in more detail, we first discuss the original Isomap algorithm.

\subsection{Isomap at a glance}
The Isomap algorithm~\cite{isomap} for finding a low-dimensional, quasi-isometric embedding of a point set $S$ consists of three steps:\smallskip

\begin{enumerate}
\item Construct a proximity graph $G$ over the point set $S$;
\item Evaluate pairwise geodesic distances between elements of $\mathcal{S}$ via Dijkstra's algorithm on $G$;
\item Perform MDS on the resulting distances to find a quasi-isometric $d$-dimensional embedding.
\end{enumerate}

\subsection{Computations involved}
\label{sec:computations}
Each algorithmic step of Isomap involves simple computations that we review next.\smallskip

\paragraph*{Proximity graph.}
A graph is first constructed by creating undirected edges between neighboring input points. Two simple ways have been proposed to define whether two points should be connected by an edge of the neighborhood graph: the first ($k$-nearest neighbor, or k-NN) approach declares two points neighbors iff one is among the $k$ nearest neighbors of the other based on Euclidean distances in $\varmathbb{R}^D$; the second ($\epsilon$-ball) approach declares two points neighbors iff the Euclidean distance between them is smaller than a user-defined threshold $\epsilon$. Both constructions can be done efficiently (typically, in $\mathcal{O}(n \log n)$) using a locality sensitive hashing data structure for instance~\changes{\cite{LHS:2004}}.\smallskip

\paragraph*{Geodesic distances.}
After setting edge weights to the Euclidean distances between corresponding pairs of points, the next step is to run Dijkstra's algorithm on the resulting weighted graph $G$ to compute approximations of all pairwise geodesic distances between points of $S$ in $\mathcal{O}(n^2 \log n)$. The squares of these pairwise geodesic distances are then assembled into an $n\!\times\! n$ symmetric matrix $\mathbf{D}$.\smallskip

\paragraph*{Multidimensional scaling.}
Finally, the (classical) MDS procedure is performed on $\mathbf{D}$ to obtain the low-dimensional embedding that preserves these squared distances as well as possible. Namely, an $n \!\times\! n$ Gram matrix $\mathbf{G}$ is constructed from $\mathbf{D}$ via ``double-centering''~\cite{MDSBook}; that is, denoting identity matrix by $\mathbf{I}$ and the $n$-vector of ones by $\mathbf{e}$, $\mathbf{G} \!=\! -\frac12 ( \mathbf{I} \!-\! \frac{1}{n} \mathbf{e\,e}^t ) \,\mathbf{D}\, ( \mathbf{I} \!-\! \frac{1}{n} \mathbf{e\,e}^t )$. (This Gram matrix may be further altered to represent a proper kernel, see~\cite{Choi:RKI:2007}).
The final embedding $\mathbf{Z}$ is found as the product of square roots of the $d$ largest eigenvalues and corresponding eigenvectors of $\mathbf{G}$:\vspace*{-2mm}
\[
\mathbf{Z} =  \sqrt{\mathbf{\Lambda}_d}\, \mathbf{Q}_d^t \vspace*{-2.5mm}
\]
where $\mathbf{G} \!=\! \mathbf{Q} \mathbf{\Lambda} \mathbf{Q}^t$ is the eigendecomposition of $\mathbf{G}$, and $\left(\mathbf{\Lambda}_d, \mathbf{Q}_d\right)$ denote the truncated matrices containing only the $d$ largest eigenvalues and their corresponding eigenvectors respectively. Consequently, the embedding is found by computing the $d$ largest eigenvalues and their associated eigenvectors, a numerical task of expected complexity $\mathcal{O}(n^3)$ since the Gram matrix is dense. MDS is, in fact, a variational approach since the solution is the minimizer over all matrices $\mathbf{Z}$ of the squared Frobenius norm $||\mathbf{Z}^t \mathbf{Z} \!-\! \mathbf{G}||^2_F$ subject to $\operatorname{rank}(\mathbf{Z}) = d$.

\begin{figure}[t]
	\includegraphics[width=\textwidth]{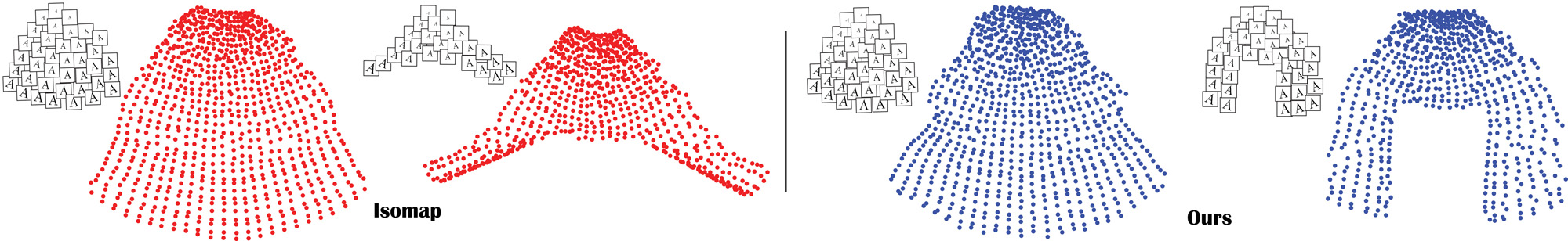}\vspace*{-3mm}
	\caption{\textbf{Letter A.}  From 888 images ($120\!\times\!120$ pixels) of rotated and resized letters `A', both Isomap (red) and PTU (blue) produce similar 2D embeddings (left, each dot indicating an image; insets show a subset of the images and their embedding). However, if a part of these input images is removed (rendering the set non-geodesically-convex), Isomap dramatically changes the embedding, while PTU properly reflects the missing images (right).\vspace*{-3mm}}
	\label{fig:letterA}
\end{figure}

\subsection{Discussion}
Despite the dense nature of the Gram matrix which implies a higher computational complexity than local methods, manifold learning through Isomap is one of the most used nonlinear dimensionality reduction methods because of its remarkable robustness to noise. Its entire reliance on graph-based shortest paths has, however, far-reaching calamitous consequences. First, spurious geodesic curvature (i.e., zigzags) in the shortest paths (seen as a piecewise linear curve in $\varmathbb{R}^D$) between two nodes on the graph  introduces inaccuracies in the estimation of geodesic distances. In practice, this issue is exacerbated by the sparse sampling that real applications often have to deal with, even if short paths can be locally rectified through a straightening projection~\cite{Max:2017}. Worse, some computational accelerations of Isomap rely on a subsampling of the initial data~\cite{lmds}, making this inaccuracy issue all the more limiting. More importantly, Isomap can only provide a quasi-isometric low-dimensional mapping for \emph{geodesically convex sets}: the presence of hole(s) or non-convex boundaries in the domain sampled by the input pointset brings significant overestimations of geodesic distances, thus distorting the results (see Fig.~\ref{fig:holyS} and Fig.~\ref{fig:letterA}). Detecting and correcting paths that go around small or large voids is difficult to perform reliably as voids can have a variety of sizes and shapes when dealing with noisy and irregular sampling.

\section{Parallel Transport Unfolding}

We now present our Parallel Transport Unfolding (PTU) algorithm, which has a very similar structure to Isomap since it comprises the following steps:\smallskip

\begin{itemize}[itemsep=0pt]
	\item Construct a proximity graph $G$ of the pointset $\mathcal{S}$ and compute local tangent spaces at each point; 
	\item Approximate all pairwise geodesic distances using parallel transport along (shortest) paths on $G$; 
	\item Perform MDS to find a $d$-dimensional embedding that best preserves all the geodesic distances.
\end{itemize}\smallskip

We detail each step next, stressing the key differences with Isomap in the first two steps of the algorithm, i.e., the use of approximate tangent spaces and of the Levi-Civita connection to better approximate intrinsic geodesic distances.

\subsection{Proximity graph}
\label{sec:graph}
The construction of a proximity graph $G$ on $\mathcal{S}$ proceeds similarly to the original Isomap algorithm: one can link each point to its neighbors contained in an $\epsilon$-ball, or to its $k$ nearest neighbors---both based on Euclidean distances in $\varmathbb{R}^D$. Variants such as a mix of the two~\cite{Wang:2006:AMM} or the mutual $k$ nearest neighbors approach~\cite{Brito:1997} can also be used to naturally discard outliers. For clarity of presentation, we will use a vanilla $k$-NN graph in our exposition and all of our tests. The value $k$ should be chosen such that the edges of the resulting graph are good approximations of geodesics between corresponding points; we will typically set $k$ around $4d$ to induce a valence greater than the usual connectivity of a regular grid of the $d$-dimensional embedding ($k\!=\!2d$), yet less than the number of $1$-ring neighbors on that same grid ($k\!=\!3^d\!-\!1$); but knowledge about noise levels in the input pointset can be employed to improve the graph quality by varying the number of neighbors from point to point. Once the graph connectivity is defined, each edge is assigned a weight equal to its Euclidean length in $\varmathbb{R}^D$ as an approximation of its geodesic length on the manifold $\mathcal{M}$.

\subsection{Tangent spaces and their orthonormal bases}
\label{sec:tangents}
For each input point $\mathbf{x}_i$, its K nearest neighbors on the proximity weighted graph (for $K\!\geq\!k$) are used to define a \emph{geodesic neighborhood}. The matrix whose rows are the data points (centered by subtracting $\mathbf{x}_i$)  in this neighborhood induces, through its $d$ left singular vectors corresponding to the $d$ largest singular values, an orthonormal basis $\mathbb{T}_{i}$ of $d$ vectors in $\varmathbb{R}^D$ spanning the approximate tangent space:\vspace*{-2mm}
\begin{equation}
\label{eq:defFrames}
\mathbb{T}_{i} = \left(\begin{matrix}
\mathbf{t}_{1}^i \quad \cdots \quad \mathbf{t}_{d}^i
\end{matrix}\right) \in \varmathbb{R}^{D \times d}.\vspace*{-2mm}
\end{equation}

While choosing $K\!=\!k$ often suffices to provide a good estimate of local tangent spaces, using a value of $K$ distinct from $k$ allows the definition of arbitrarily large geodesic neighborhoods (useful in the presence of strong noise) around each point of $\mathcal{S}$, while alleviating the traditional issue of ``manifold shortcutting'' associated with increasing $k$~\cite{Max:2017}.
As a substitute to this frame construction via partial SVD, note that an improved approximation of the local tangent spaces in the presence of strong noise and outliers can also be computed via $\ell_1$-based robust PCA (see, e.g.,~\cite{Zhang:2014:NMR}).

With these orthonormal frames of tangent spaces in place,  we can now discuss how to approach discrete parallel transport and how to use it for geodesic length estimation.

\subsection{Discrete Parallel transport}

We now cover the core of our approach, i.e., exploiting parallel transport in high dimension to better evaluate geodesic distances.\smallskip

\paragraph*{Parallel Transport in Differential Geometry.}
The notion of parallel transport plays a central role in differential geometry. It induces a way to connect the geometries of nearby points, thus prescribing how a basis of the tangent space at one point of a manifold should be adjusted to produce a ``parallel'' basis of another tangent space at a nearby point. While this procedure is straightforward for flat spaces (it corresponds to a simple translation), it becomes more involved for manifolds with non-trivial curvature. Its differential geometric treatment involves the definition of a connection on the tangent bundle~\cite{Spivak:1979, Frankel:2011}, which represents an infinitesimal analogue of parallel transport. Most relevant to our work are \emph{metric} connections, i.e., connections such that the parallel transport they define preserves the intrinsic metric of the manifold. The well-known Christoffel symbols are, in fact, the components of a particularly canonical metric connection called the Levi-Civita connection, which we will leverage in our application. Another geometric property we will exploit is the fact that geodesics, usually described through variational analysis as locally shortest curves, can also be defined through parallel transport: a geodesic is a curve that parallel transports its own tangent vector \changes{as directly implied by the geodesic equation~\cite{Spivak:1979,Frankel:2011}}. This simple property will guide our evaluation of geodesic distances.\smallskip

\paragraph*{Discrete Parallel Transport.}
Given points $\mathbf{x}_i$ and $\mathbf{x}_j$ sharing an edge in the proximity graph $G$, we define the \emph{discrete metric connection} between $\mathbf{x}_i$ and $\mathbf{x}_j$ as the orthogonal $d\times d$ matrix $\mathbf{R}_{j,i}$ in $O(d)$ representing the change of basis that best aligns, in the Frobenius norm, the frames $\mathbb{T}_i$ and $\mathbb{T}_j$, i.e.,\vspace*{-2mm}
\begin{equation}
\mathbf{R}_{j,i} =\argmin_{\mathbf{R} \in O(d)}  ||\mathbb{T}_i - \mathbb{T}_j \mathbf{R}||_F^2 \vspace*{-2mm}
\label{eq:defRij}
\end{equation}
By definition, $\mathbf{R}_{i,j}$ is the inverse of $\mathbf{R}_{j,i}$: $\mathbf{R}_{i,j} \!=\! \mathbf{R}_{j,i} \,\!\!\!\!\!^t\;$.
Note that we use the group $O(d)$ of orthogonal matrices because the SVD used in Sec.~\ref{sec:tangents} produces arbitrarily-oriented tangent frames. A pre-processing of these tangent bases could be performed (e.g., via a minimum spanning tree) if one wants to ensure that their orientations are consistent, enforcing that the discrete connections are pure rotations (in $S\!O(d)$) as in the continuous case. The transformation $\mathbf{R}_{j,i}$ can thus be understood as a discrete equivalent to the Levi-Civita connection induced by the metric on $\mathcal{S}$ inherited from the Euclidean space $\varmathbb{R}^D$. It is also a high-dimensional extension of previous discretization of metric connections on triangle meshes~\cite{Crane:2010:TC,Liu:2016:DCC}.\\[-4mm]

Computing this discrete connection is easily achieved via \changes{a singular value decomposition}:\smallskip

\noindent\emph{\textbf{Proposition \changes{1}: }
Let $\mathbb{T}_i\ \!\!^t \mathbb{T}_{\!j} = \mathbf{U} \mathbf{\Sigma} \mathbf{V}^t$ be the \changes{SVD}. The discrete connection is expressed as:}\vspace*{-2mm}
\begin{equation}
\label{eq:defConnection}
\mathbf{R}_{j,i}=\mathbf{V} \mathbf{U}^t\!.\vspace*{-1mm}
\end{equation}

\noindent\emph{Proof: }
The definition of $\mathbf{R}_{i,j}$ in Eq.~(\ref{eq:defRij}) maximizes $\operatorname{Tr}\!\left[ \mathbb{T}_i\ \!\!^t \mathbb{T}_{\!j} \mathbf{R} \right]$ for $\mathbf{R}\!\in\!O(d)$ as the other terms do not depend on $\mathbf{R}$. Since $\mathbb{T}_i\ \!\!^t \mathbb{T}_{\!j} \mathbf{R}  \!=\! \mathbf{U} \mathbf{\Sigma} \mathbf{V}^t \mathbf{R}$ and given the invariance of the trace under cyclic permutations, the optimal rotation maximizes $\operatorname{Tr}\!\left[ \left(\mathbf{V}^t \mathbf{R} \mathbf{U}\right) \mathbf{\Sigma} \right]$ for $\mathbf{V}^t \mathbf{R} \mathbf{U} \!\in\! O(d)$. Consequently, one has $\mathbf{V}^t \mathbf{R} \mathbf{U} = \mathbf{Id}$; so $\mathbf{R}_{j,i}$ must be $\mathbf{V} \mathbf{U}^t$. One recognizes the Procrustes superimposition of two nearby frames~\cite{Kabsch:BestRot:1976}, just extended to handle arbitrary choices of frame orientation for tangent spaces. \hfill\ensuremath\blacksquare
\vspace*{2mm}

\subsection{Local Path Unfolding via Parallel Transport}
\label{sec:unfolding}

With a discrete metric connection, we can now ``unfold'' a polyline path on $\mathcal{S}$ (made out of a series of adjacent proximity graph edges) into $\varmathbb{R}^d$, i.e., we seek to \emph{map a polyline of graph edges into a flat $d$-dimensional space while best preserving its metric properties such as length, intrinsic curvatures, etc}. This process of ``unrolling'' a curve onto a tangent space is known as Cartan's development~\cite{Nomizu:1978,Sharpe:1997} in differential geometry.\smallskip

\paragraph*{Two-edge Unfolding.}
Let us define how to unfold in $\varmathbb{R}^d$ a three-point ($\mathbf{x}_i,\mathbf{x}_j,\mathbf{x}_k$) polyline path on $\mathcal{M}$, i.e., a path made of two adjacent edges of the proximity graph on $\mathcal{S}.$ Its extensions to arbitrary paths will be straightforward. Without loss of generality, we can consider the unfolding to be happening in the tangent space at $\mathbf{x}_i$, equipped with its orthonormal frame $\mathbb{T}_i$. The first point $\mathbf{y}_i$ of the unfolded polyline can be chosen to be at the origin of this subspace, or coinciding with $\mathbf{x}_i$. The first edge, represented by the vector $\mathbf{e}_i\!=\!\mathbf{x}_j \!-\! \mathbf{x}_i$, is projected (in the $\ell^2$ sense) onto the tangent space at $\mathbf{x}_i$ to form a $d$-dimensional tangent vector $\mathbf{v}_i$ through:\vspace*{-2mm}
\[
\mathbf{v}_i = \mathbb{T}_i^t \mathbf{e}_i.\vspace*{-2mm}
\]
This vector $\mathbf{v}_i$ is then added to $\mathbf{y}_i$ to form the point $\mathbf{y}_j$, thus defining the first unfolded edge. The second edge, $\mathbf{e}_j\!=\!\mathbf{x}_k \!-\! \mathbf{x}_j$ is similarly projected onto the local tangent space defined by frame $\mathbb{T}_j$ at $\mathbf{x}_j$. The resulting vector can then be parallel transported onto the tangent space at $\mathbf{x}_i$ to become:\vspace*{-2mm}
\[
\mathbf{v}_j = \mathbf{R}_{i,j} \left[ \mathbb{T}_j^t \mathbf{e}_j \right],\vspace*{-2mm}
\]
where the term in brackets is the $\ell^2$ projection of $\mathbf{e}_j$ onto $\mathbb{T}_j$. From this vector, now represented in the original tangent frame at $\mathbf{x}_i$, we construct the final point of the unfolded polyline as: $\mathbf{y}_k \!=\! \mathbf{y}_j + \mathbf{v}_j$.\smallskip

\paragraph*{Preservation of geodesic curvature.}
Due to our use of a discrete metric connection, the unfolding procedure we described has an important property: it nearly preserves the geodesic curvature of the initial curve. Indeed, we used a discretization of the metric-preserving Levi-Civita connection, so the \emph{intrinsic} angle between vectors $\mathbf{e}_i$ and $\mathbf{e}_j$ (meaning, the angle measured on the manifold $\mathcal{S}$) is preserved by parallel transport, and corresponds to the angle between $(\mathbf{y}_i,\mathbf{y}_j)$ and $(\mathbf{y}_j,\mathbf{y}_k)$ in $\varmathbb{R}^d$---up to discretization errors. This means that if the two-edge polyline were a good approximation of a geodesic on $\mathcal{S}$, the unfolded polyline would be (nearly) straight in $\varmathbb{R}^d$ since a geodesic curve parallel transports its tangent vector. As a corollary, the Euclidean distance $\| \mathbf{v}_i + \mathbf{v}_j\|_2$ between $\mathbf{y}_i$ and $\mathbf{y}_k$ is a good approximation of the geodesic distance between $\mathbf{x}_i$ and $\mathbf{x}_k$ along $\mathcal{S}$---whether the polyline is an approximate geodesic or not.\smallskip

\paragraph*{Length rescaling.}
In case of highly-irregular samples with small or no noise, it can be beneficial to preserve the length of each edge $\mathbf{e}_i$ once it is projected onto the corresponding frame, as the projection $\mathbf{v}_i$ on the tangent space can be shortened if the manifold is curved. This adjustment is easily achieved by, for instance, setting\vspace*{-1.5mm}
\[
\mathbf{v}_i = \mathbb{T}_i^t \mathbf{e}_i \frac{\|\mathbf{e}_i\|_2}{\| \mathbb{T}_i^t \mathbf{e}_i\|_2}.\vspace*{-1.5mm}
\label{eq::lengthRescaling}
\]
If strong noise and/or outliers are present, this modification should be avoided as it tends to overestimate local distances around noisy points. We only used this length rescaling procedure in Figs.~\ref{fig:letterA} and~\ref{fig:irregularS} to slightly improve the results in these extreme cases.

\subsection{Distance approximation via Parallel Transport}
\label{sec:ptu_dist}

The unfolding procedure we just introduced can be used to evaluate pairwise geodesic distances without the traditional shortcomings of Dijkstra's algorithm for graphs over an irregular sampling.\smallskip

\begin{figure}[t]
\centering
	\includegraphics[width=0.9\textwidth]{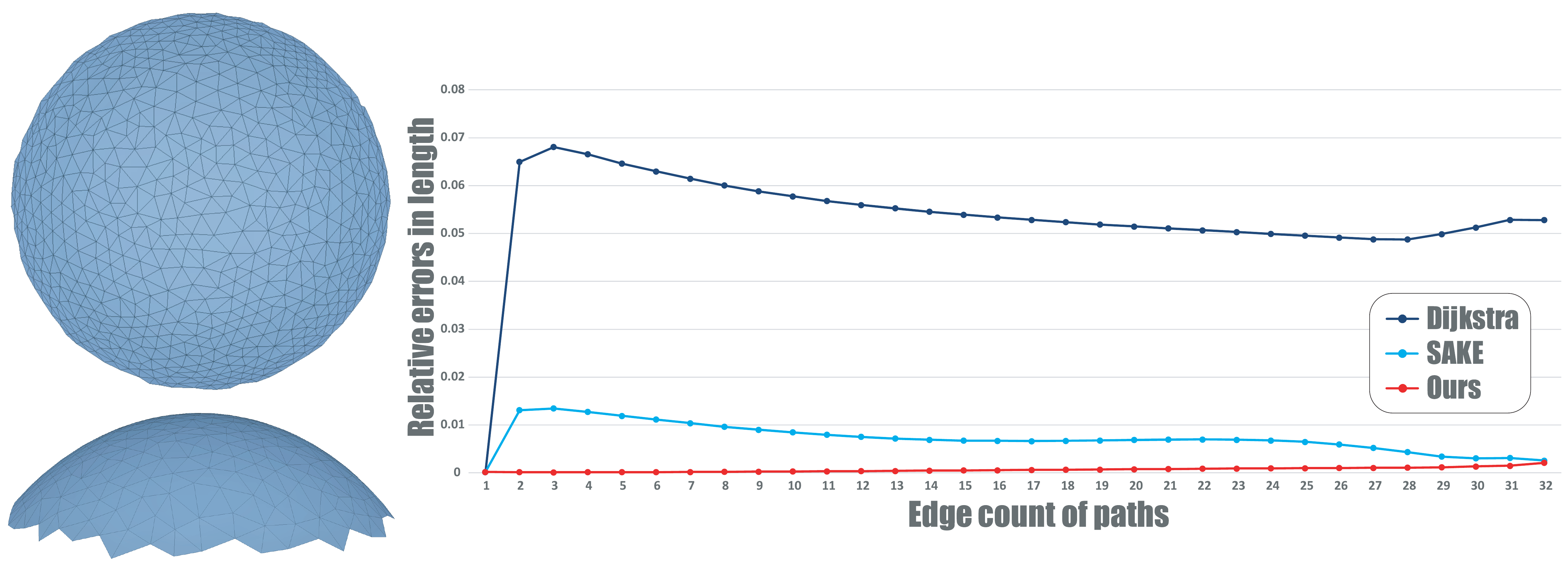}\vspace*{-3mm}
	\caption{\textbf{Geodesic Distances.}
		For a graph based on a triangulated spherical cap (left, 2 views), the average relative error of pairwise Dijkstra-based geodesic distances is 5.6\%; SAKE~\protect\cite{Max:2017} already reduces the error to 0.8\%, but PTU brings it down to 0.046\%. Plots (right) show relative errors of pairwise distances as a function of their number of edges.\vspace*{-3mm}}
		\label{fig:improvedGeodesics}
\end{figure}

\paragraph*{Correcting shortest paths.}
Consider a Dijkstra shortest polyline $\left(\mathbf{x}_{i_1},..., \mathbf{x}_{i_m}\right)$. 
Using the unfolding procedure into $\mathbb{T}_{i_1}$, we iteratively project the edges
$\mathbf{e}_{i_s} \!=\! \mathbf{x}_{i_{s+1}} - \mathbf{x}_{i_s}$ onto the tangent space spanned by $\mathbb{T}_{i_s}$, before parallel transporting the resulting vector back to the original tangent space, i.e.,\vspace*{-2mm}
\begin{equation}
\mathbf{v}_{i_s}= \biggl(\prod_{j=1..r-1} \mathbf{R}_{i_j,i_{j+1}} \biggr) \biggl[\mathbb{T}^{\ t}_{i_{r}} \mathbf{e}_{i_s} \biggr].\vspace*{-2mm}
\label{eq::geodFormula}
\end{equation}

After accumulating the results into a single vector $\mathbf{v} \!=\! \sum_{s=1..m} \mathbf{v}_{i_{s}}$, we set the approximate geodesic distance between $\mathbf{x}_{i_1}$ and $\mathbf{x}_{i_m}$ to be its length $||\mathbf{v}||_2$: as discussed above, the span of the unfolded path is a better approximation of the geodesic distance between the end points of the initial path as it ignores the intrinsic twists and turns that the path went through.
Since our estimate of the geodesic distance from $\mathbf{x}_p$ to $\mathbf{x}_q$ is in general not the same as the estimate from $\mathbf{x}_q$ to $\mathbf{x}_p$ due to the asymmetry of the discrete parallel transport, we average the two spans in a final post-processing step to determine all the pairwise geodesic lengths.
Fig.~\ref{fig:improvedGeodesics} demonstrates an improvement of \emph{over two orders of magnitude} compared to Dijkstra's approach to estimating geodesic distances.\smallskip

\paragraph*{Leveraging unfolded paths.} While we only need the Euclidean distance between the two end points of the unfolded path in $\varmathbb{R}^d$ to perform nonlinear manifold learning, the unfolded path can also be useful for gathering additional information. For instance, the largest distance between the straight line between the two end points and the actual unfolded path indicates how far off the path is from being a geodesic. A large value is a typical telltale of the presence of a large void in the data within the manifold, which can be exploited to suggest where to insert new samples in order to improve the results. Similarly, a large difference between the distance estimates computed from one end point to the other and in the opposite direction implies potential issues with sampling density of the data compared to the local curvature of the manifold.\smallskip

\paragraph*{Connection-based Dijkstra's algorithm.}
Finally, we point out that our parallel transport approach to estimating geodesic
distances could be done along \emph{approximately-shortest} polyline between a given pair of points: fast approximations of shortest paths could thus be used to lower this $\mathcal{O}(n^2 \log n)$ step without significant effect on the results. However, the longest the polyline, the more likely numerical inaccuracies induced by repeated alignment of frame fields will accumulate. Since Dijkstra's algorithm is not the computational bottleneck in Isomap, we decided to utilize graph-based shortest paths: in fact, our construction can be neatly incorporated within the dynamic programming approach that Dijkstra's algorithm uses---i.e., storing the predecessor to each point in the shortest path found thus far. Parallel transport unfolding only requires the insertion of six lines in the original Dijkstra's algorithm in order to reduce the unnecessarily repeated calculations that result from the fact that sub-paths of shortest paths are also shortest paths. Specifically, every time a point is removed from the priority queue (i.e., every time the algorithm finds a shorter path to a point), we can compute the corresponding vector $\mathbf{v}_i$, before storing the cumulative connection (i.e., the product of connection matrices along the path) to be the cumulative connection of the predecessor multiplied by the local discrete connection. Proceeding in this way guarantees that the calculations are performed without redundancies, adding a negligible amount of computational time to the traditional Dijkstra's algorithm. Pseudocode is given in Alg.~\ref{alg:ptu_dijkstra}.

\begin{figure}[t]
	\includegraphics[width=\linewidth]{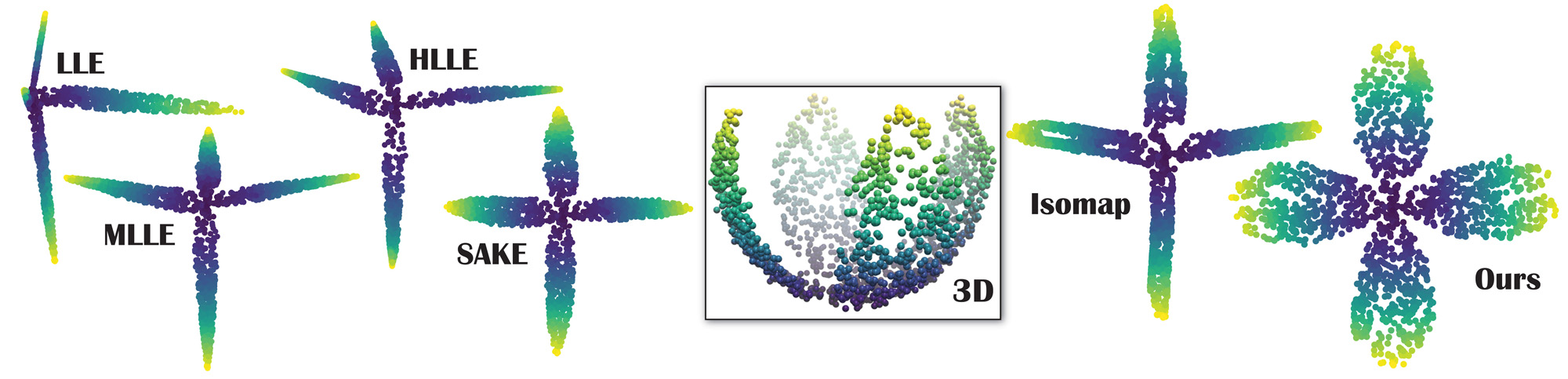}\vspace*{-4mm}
	\caption{\textbf{Noisy Petals.} Given a 3D sampling of a 4-petal shaped portion of a sphere (see Fig.\ref{fig:Petals}) with added Gaussian noise in the normal direction ($\sigma$: 3\% of sphere radius, middle), PTU recovers an almost perfect quasi-isometric 2D parametrization, while Isomap still fails (right). Local methods, not exploiting large geodesic distances, fail even worse (left) with the notable exception of SAKE that performs better than Isomap. \vspace*{-4mm}}
	\label{fig:Petals2}
\end{figure}

\subsection{MDS}
Finally, once we evaluated all pairwise geodesic distances, the results can be processed through MDS: double-centering the matrix of \textit{squared} distances produces the Gram matrix, whose partial eigendecomposition returns the final embedding as in Sec.~\ref{sec:computations}. Numerical improvements (see~\cite{Harel:2002,Brandes:2007,Halko:2011,Khoury:2012}) can be applied to reduce the cubic complexity of this stage. \changes{Note that robust alternatives to classical MDS could also be used here (e.g.,~\cite{Cayton:2006, Kovnatsky:2016}) to offer resilience to outliers, but at higher computational cost.}

\subsection{Discussion}
We conclude this section with a few properties worth mentioning.\\[-3mm]

\noindent \emph{\textbf{Proposition \changes{2}:} Unlike Isomap, PTU is linearly precise as long as one uses a proximity graph $G$ where each sample point has enough neighbors to span a $d$ dimensional subspace: assuming the pointset $\mathcal{S}$ samples a linear $d$-dimensional subspace $\mathcal{R}$ of $\mathbb{R}^D$, the PTU embedding $\mathbf{Z}$ is isometric to $\mathbf{X}$.} \\[-3mm]

\emph{Proof:}  When the sampled data lie on a linear subspace of dimension $d$, each orthonormal frame $\mathbb{T}_{i}$, computed using SVD as described in Sec.\ref{sec:ptu_dist}, forms a basis for $\mathcal{R}$. Moreover, every pair $\mathbb{T}_{i}$ and $\mathbb{T}_{j}$ are perfectly aligned by the discrete connection $\mathbf{R}_{j,i}$ (Eq.\eqref{eq:defRij}). As a result, unfolding a polyline $\left(\mathbf{x}_{i_1},..., \mathbf{x}_{i_m}\right)$ reduces to rewriting it in the basis $\mathbb{T}_{i_1}$ and our geodesic length estimation recovers the exact (Euclidean) distance between $\mathbf{x}_{i_1}$ and $\mathbf{x}_{i_m}$, independent of the sampling irregularities or of the geodesic convexity of the domain. PTU thus becomes equivalent to classical MDS, producing a $d$-dimensional embedding $\mathbf{Z}$ that is isometric to $\mathbf{X}$ by construction.\hfill\ensuremath\blacksquare\\[-3mm]

\noindent \emph{\textbf{Proposition \changes{3}:}
Under mild assumptions on the regularity of a manifold $\mathcal{M}$ and assuming  that the input pointset $\mathcal{S}$ samples $\mathcal{M}$ finely enough with potential sampling voids over regions of small (sectional) curvature of the manifold, the PTU estimate $d_\text{\tiny PTU}(\mathbf{x}_{i},\mathbf{x}_{j})$ of the geodesic distance between points $\mathbf{x}_{i}$ and $\mathbf{x}_{j}$ based on a Dijkstra (shortest) polyline $(\mathbf{x}_{i},..., \mathbf{x}_{j})$ computed on the proximity graph $G$ of $\mathcal{S}$ approximates the real geodesic distance $d_{g}(\mathbf{x}_{i},\mathbf{x}_{j})$.}
\\[-3.5mm]

\emph{Discussion:} We quantify this proposition more rigorously in App.~\ref{app:proof} by providing a concrete error bound (and its proof) between $d_\text{\tiny PTU}$ and $d_g$. Our bound relies on three key components: 1) if the pointset $\mathcal{S}$ is dense enough, then for appropriate choice of parameters $k$ and $K$ the discrete tangent spaces approximate their continuous equivalents, and the resulting discrete connection converges to the Levi-Civita connection in the sampling limit as proved in~\cite{Singer:2012}, where the authors used an equivalent notion of parallel transport to define and study a discrete connection Laplacian; 2) sampling voids in $\mathcal{S}$ are allowed as long as the integral of the intrinsic sectional curvature of $\mathcal{M}$ over the regions of the underlying manifold corresponding to these voids is small---in other words, we will assume that a Dijkstra polyline lies within a tubular neighborhood of its corresponding geodesic, for a tubular diameter less than $\mathcal{O}(1/\!\sqrt{\kappa_s})$ where $\kappa_s$ is the local maximum absolute value of the sectional curvature of $\mathcal{M}$; and 3) for a dense enough sampling of $\mathcal{M}$, a straight $\varmathbb{R}^n$ vector between two nearby points $\mathbf{x}_i$ and $\mathbf{x}_j$ on $\mathcal{M}$ has approximately the same a length as the Cartan development of the geodesic between them on the tangent space $T_i\mathcal{M}$ at $\mathbf{x}_i$. Note that our statement involves no geodesic convexity requirement. 
	
The benefit of using parallel transport over regular Dijkstra geodesic length approximation is thus clear: our construction eliminates spurious geodesic curvature that graph approximations inevitably suffer from, bringing significant improvement even for well sampled domains (see Fig.~\ref{fig:improvedGeodesics}). It also allows for almost perfect recovery of pairwise geodesic distances for developable manifolds ($\kappa_s\!=\!0$) with \emph{arbitrary} topology in the sampling limit, even if the sampled manifold is not geodesically convex. When voids are present in pointsets that sample non-developable manifolds, PTU computes approximate geodesic distances without having to explicitly fill in the voids. Instead, it extracts geometric information from paths surrounding each hole to recover high accuracy geodesic estimates on the manifold, provided that the voids were not over regions with large curvature.\\[-3mm]

\noindent \emph{\textbf{Proposition \changes{4}}: Just like Isomap, the complexity of PTU is $O(n^3)$.}\\[-3.5mm]

\emph{Proof:} Our approach does not change the computational complexity of the various steps compared to Isomap: the proximity graph construction is still $O(n \log n)$, the construction of tangent planes takes $O(n\,D\,K^2),$ our Parallel Transport Unfolding using Dijkstra shortest paths is in $O(n^2 \log n)$ (the unfolding process part itself requires $O(n^2 (Dd^2 + d^3)$ to perform matrix multiplications and SVDs), while the partial eigendecomposition of a dense matrix is expected to take $O(n^3)$ operations.\hfill\ensuremath\blacksquare\\[-2.5mm]

\begin{figure}[t]
	\centering
	\includegraphics[width=.95\linewidth]{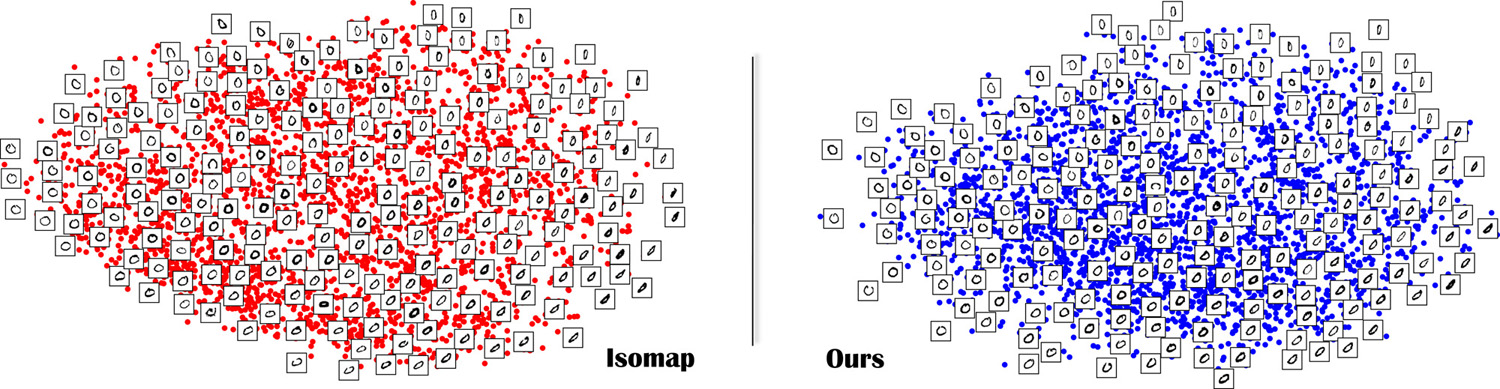}\vspace*{-3mm}
	\caption{\textbf{Digit zero.} The 3000 images of handwritten zeros (with a resolution of 28x28 pixels) from the MNIST dataset are mapped quite similarly by both Isomap and PTU for the first two most significant coordinates (here, $d\!=\!4$ was used). A few images are displayed next to their corresponding 2D point for visualization purposes.\vspace*{-3mm}}
	\label{fig:Digit0}
\end{figure}

\paragraph*{Additional control.} Compared to Isomap, our parallel transport based approach has a few extra parameters that can be exploited to offer more control over the manifold learning process: \vspace*{-1mm}
\begin{itemize}[itemsep=0pt]
\item Since we compute local tangent spaces at each input point, the neighborhood size $K$ can be adapted (either globally or locally) to the input. While using $K\!=\!k$ is sufficient in most cases (see our examples), raising this value can help deal with very noisy inputs as mentioned in Sec.~\ref{sec:tangents}.
\item Similarly, local PCA of these neighborhoods may not lead to the best estimations of tangent spaces in extreme cases: robust PCA, or even local averaging of the PCA estimates\footnote{As a side note, having a discrete connection makes the local averaging of vectors or frames particularly simple, as neighboring values can be parallel transported to a common point, where a pointwise average is computed.} can help manage large amounts of noise and outliers.
Again, we did not have recourse to these variants to prove the robustness of PTU in its default form; but they can be easily incorporated in a practical implementation of our approach to add flexibility. 
\item Finally, our parallel transport estimation of geodesic lengths assumes the knowledge of the dimension $d$ of the data, just like Isomap requires as well---and many approaches have been proposed to estimate this dimension directly from the data, see, e.g.,~\cite{Pettis:1979:IDE,Levina:2004:MAX}. Yet, this intrinsic dimension can, in fact, differ from the dimensionality of the visualization one wishes to produce. Fig.~\ref{fig:Digit0} illustrates this point: dimensionality reduction approaches applied on the MNIST dataset of digits often use a 2D illustration of their results for easy visual display; however, local analysis of the dimensionality of the zero digit image set indicates an intrinsic dimension of $d\!=\!4$, reflecting the variety of ways to pen a zero (slant, thickness, smoothness, ...). While this information cannot be exploited in Isomap if a 2D depiction is desired, PTU can exploit this estimate of $d$ for its parallel transport procedure, but use only the first two eigenvectors of the Gram matrix---essentially showing a 2D projection of a 4D parameterization of the dataset. Fig.~\ref{fig:Faces} shows another example of a 2D visualization of an intrinsic parameterization for $d\!=\!3$. 
\end{itemize}

\paragraph*{Related geometric methods.}
Note finally that two related works have proposed using parallel transport for data analysis, albeit for different purposes. Vector diffusion maps (VDM~\cite{Singer:2012}) also exploit parallel transport, but focus instead on computing a low-dimensional embedding that preserves \emph{vector diffusion distances} derived from a connection Laplacian, not true geodesics (see their Figs. 6.2 \& 6.3).
Parallel vector field embedding (PVF~\cite{Lin:2013}) proposes a different discretization of the connection-Laplacian relying on an extrinsic definition of the covariant derivative. Coordinates of an embedding are constructed via Poisson solves, so that the coordinate lines in $\varmathbb{R}^d$ are mapped to parallel vector fields on the original data. Just like our approach, PVF perfectly recovers an isometric parametrization if the manifold is developable, since it is then isometric to a subset of $\varmathbb{R}^d$. However, for non-developable manifolds, coordinate lines of the resulting parameterization do \emph{not} correspond to geodesic curves on the original manifold. While they provide a valid definition of \emph{an} embedding, PVF look for a low-dimensional ``quasi-parallel'' embedding. PTU, instead, targets the same goal as Isomap, i.e., a quasi-isometric mapping for arbitrary sampled manifolds.

\subsection{Acceleration via landmarks}
A particularly simple way to accelerate Isomap is through the use of ``landmarks''~\cite{l_isomap}, a small fraction of samples of the original pointset: the MDS procedure is applied just to the landmarks to find their quasi-isometric embedding $\mathbbm{Z}$ in $\varmathbb{R}^d$, before positioning all other points \emph{relative to those landmarks} in linear time, significantly reducing computational complexity. This approach, however, often fails in practice: the sensitivity of the original Isomap method to poor sampling quality makes the low-dimensional embedding of a few landmarks very brittle.\smallskip

Given its much greater robustness to irregular sampling, PTU is particularly amenable to this landmark-based acceleration without any other alteration than replacing distance estimation by our parallel transport approach. If $\ell$ landmarks are used, the amount of computations can decrease quite dramatically: the MDS complexity (which was the bottleneck for Isomap and PTU) changes from $\mathcal{O}(n^3)$ to $\mathcal{O}(\ell^3)$. Let us briefly discuss the implementation of such an L-PTU variant.\smallskip

\paragraph*{Setup.} From the $n$ original points, we extract $\ell$ landmarks with $\ell\!\ll\!n$, used as a coarse approximation of the input geometry. Landmark selection is not a sensitive part of our approach as long as the landmarks provide a good spatial coverage of the initial pointset. Note also that we compute our discrete connection for the entire pointset, since 1) it is not the computational bottleneck of the original PTU treatment, and 2) we need the distance from every data point to every landmark in order to compute the final embedding anyway. The choice of computing a ``full resolution'' connection guarantees accurate estimation of geodesic distances even if very few landmarks are used.\smallskip

\begin{figure}[t]
	\subfloat[9 and 19 landmarks on Petals]{\includegraphics[width=0.37\linewidth]{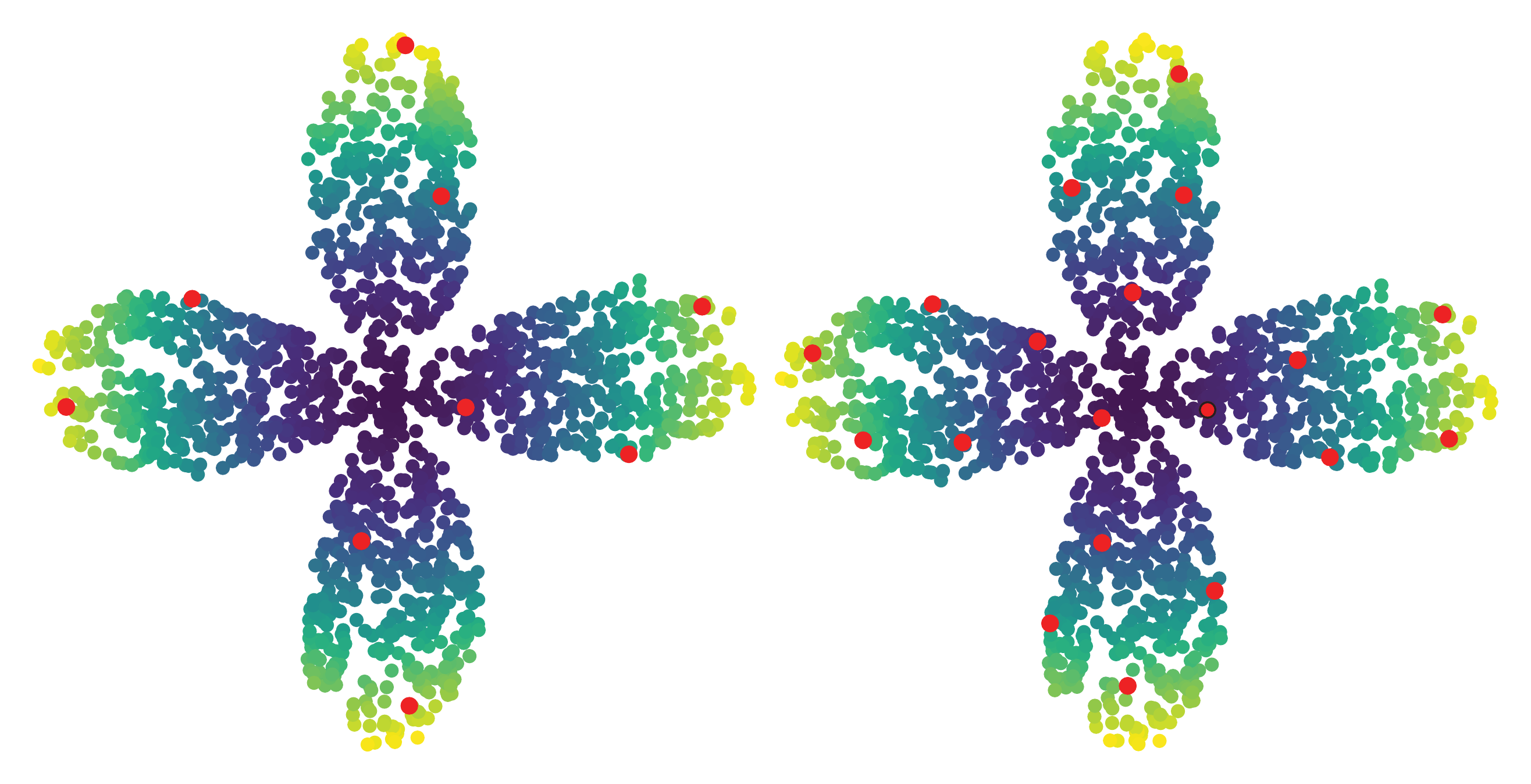}}\hfill \vrule\hfill
	\subfloat[5, 10 and 15 landmarks on letter A]{\includegraphics[width=0.6\linewidth]{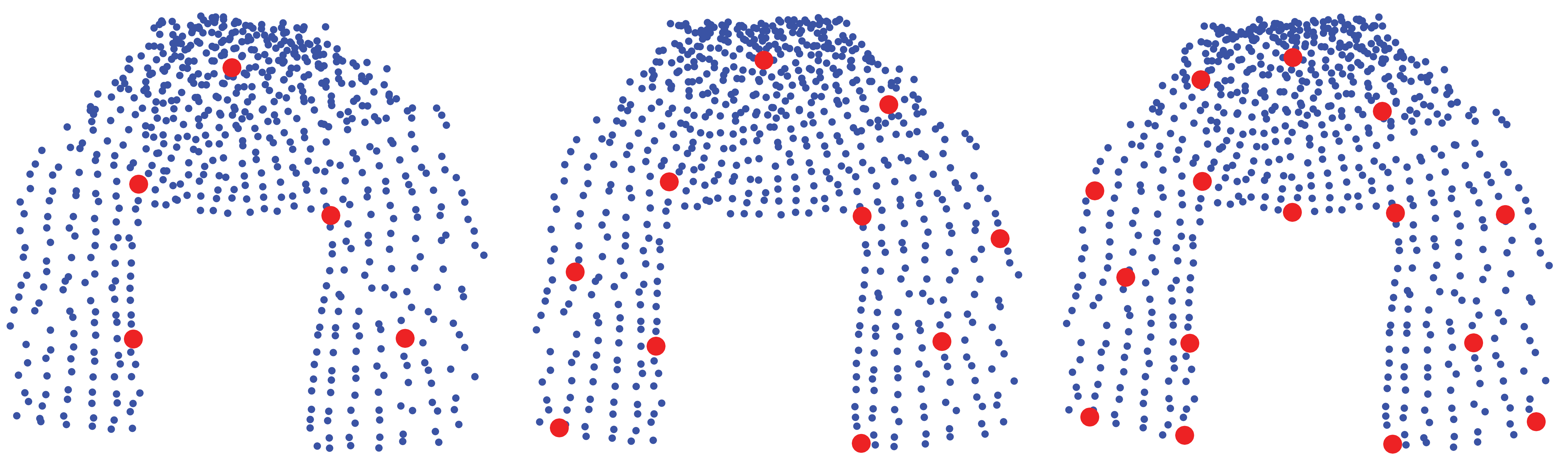}}\vspace*{-1mm}
	\caption{\textbf{Landmark-PTU.} Combining parallel transport unfolding with the landmark-based approach of L-Isomap~\protect\cite{l_isomap} drastically reduces computational times, with only small differences in the embedding: (top) for 9 (left) and 19 (right) landmarks, results on the Noisy Petals dataset are visually indistinguishable from the full treatment (see Fig.~\ref{fig:Petals2}); (bottom) for the Letter A dataset, 5 landmarks (left) already capture the proper embedding, but 10 (middle) and 15 (right) landmarks result in a better approximation of the full treatment found in Fig.~\ref{fig:letterA} (landmarks are in red). \vspace*{-2mm}}
	\label{fig:L-PTU}
\end{figure}

\paragraph*{From landmark embedding to pointset embedding.}
After computing a low-dimensional embedding of the $\ell$ landmarks using PTU as described in Sec.~\ref{sec:computations} through \vspace*{-1mm}
\[\mathbbm{Z} = \sqrt{\mathbbold{\Lambda}_{d}}\, \mathbbm{Q}_{d}^t,\vspace*{-1.2mm}\]
(where now $\mathbbold{\Lambda}_d$ and $\mathbbm{Q}_d$ are the $d$ largest eigenvalues stored in a diagonal matrix and corresponding eigenvectors of the Gram matrix derived from the $\ell \times \ell$ matrix $\lD$ of the \emph{squared} geodesic distances between landmarks only),
the embedding of the remaining points can be performed using its pseudoinverse $\mathbbm{Z}^\dagger$ and the knowledge of geodesic distances between input points and landmarks. The pseudoinverse of $\mathbbm{Z}$ has an explicit formulation due to its basic form, requiring no additional computations: \vspace*{-1mm}
\[\mathbbm{Z}^\dagger =  \mathbbm{Q}_{d} \sqrt{\mathbbold{\Lambda}_{d}}^{-1} \vspace*{-1.5mm}\]
Denote by $\avld$ the columnwise mean of $\lD$ (i.e., $\avld_p \!=\! \tfrac1\ell\sum_q \lD_{pq}$). Then as described in L-Isomap~\cite{l_isomap}, the position of a non-landmark point $\mathbf{x}_i \!\in\! \mathcal{S}$ in the low-dimensional embedding can be directly computed using a vector $\ld_i$ of squared geodesic distances from $\mathbf{x}_i$ to the $\ell$ landmarks through:\vspace*{-1mm}
\[ \mathbf{z}_i = \bigl(\mathbbm{Z}^\dagger\bigr) ^t \left(\, \avld - \ld_i \right) \vspace*{-1.5mm}\]
That is, knowing how distant a point is from the landmarks originally, we deduce its final position based on the low-dimensional embedding of the landmarks in $\mathcal{O}(\ell^2)$. Consequently, the proposed L-PTU procedure (consisting of computing all pairwise geodesic distances between $\ell$ landmarks and all input points, constructing embedding of the landmarks via MDS, and enriching it with non-landmarks through matrix-vector multiplications) brings the complexity down to $\mathcal{O}(\ell n \log (n) + \ell n (Dd^2 + d^3) + \ell^3 + \ell^2 n)$.  As Fig.~\ref{fig:L-PTU} demonstrates, this simple variant returns nearly the same embeddings as full PTU with as little as 0.1\% to 0.5\% of the points used as landmarks. More results using L-PTU are available in the Supplementary Material.

\begin{figure}[t]
\centering
\subfloat[3D torus in 4D]{\label{fig:SimpleTorus}	\includegraphics[width=0.675\linewidth]{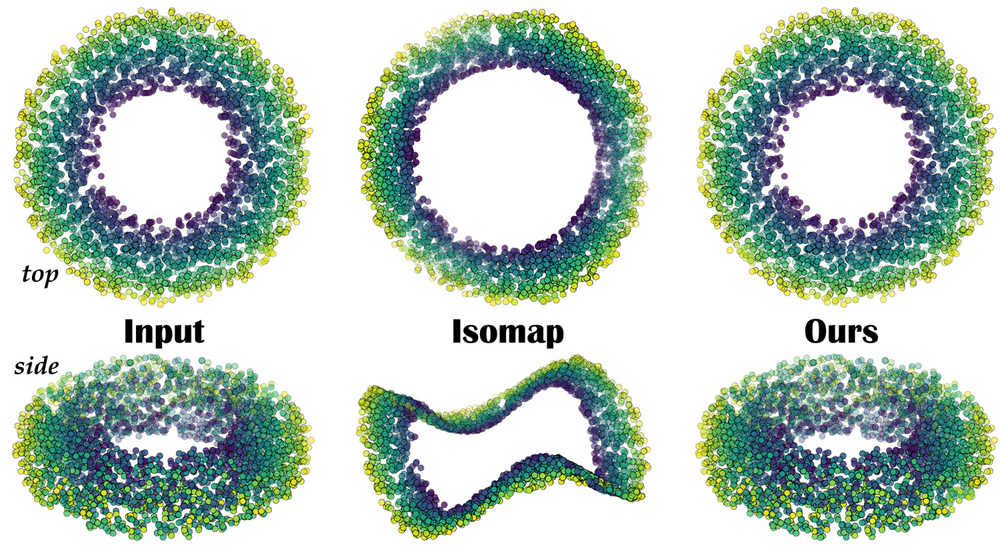}}
\quad\vrule\quad
\subfloat[Curved torus in 4D]{\label{fig:CurvedTorus}	\includegraphics[width=0.225\linewidth]{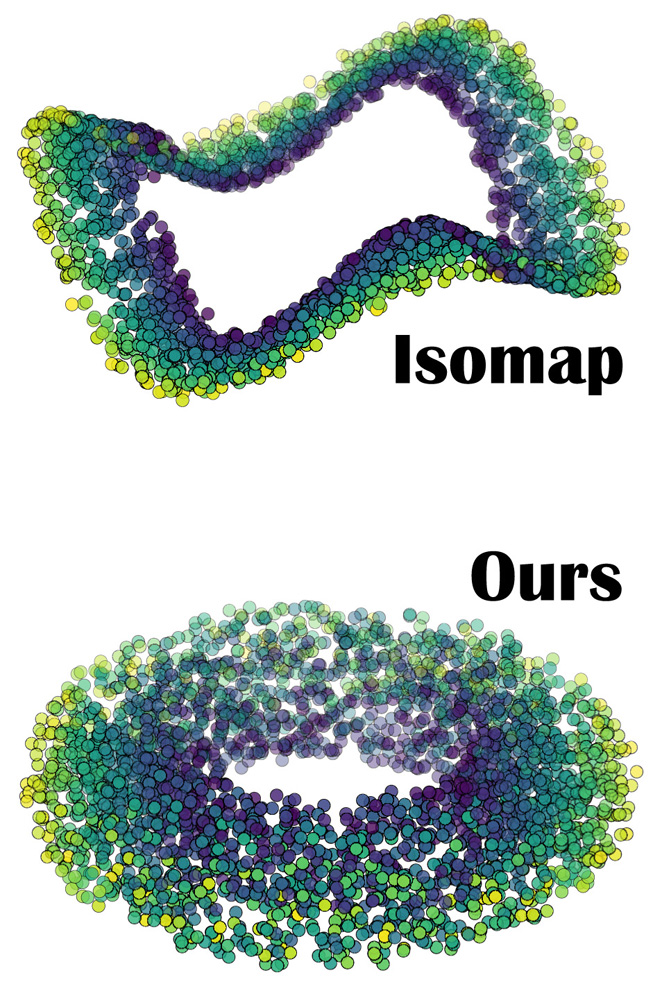}}\vspace*{-2mm}
	\caption{\textbf{Tori embedded in 4D.} (a) From a 3D pointset filling up a toroidal domain that we trivially embedded in 4D by adding an extra constant coordinate, the 3D embedding computed via Isomap leads to an unexpected global distortion due to the manifold being not convex. Instead, PTU is nearly perfect: Isomap has a normalized average relative error of $40.6\%$, while PTU is $0.06\%$. (b) A mildly curved torus in 4D is obtained by mapping a set of points $(x,y,z) \in T^2$ to $(x,y,z, (x^2 + y^2)/2)$. Now it has non-zero curvature as a manifold in 4D.
3D Isomap embedding still suffers from similar global distortion (left); PTU recovers the torus very well (right) despite the non-trivial curvature. \vspace*{-2mm}}
\end{figure}

\section{Results}
\label{sec:results}
We now discuss implementation details and provide a series of tests, on synthetic and real datasets, to compare our parallel transport approach to the original Isomap and other nonlinear dinensionality reduction methods.

\subsection{Implementation details}
Our implementation follows the steps described in the previous sections as summarized in Alg.~\ref{alg:ptu}. We use a modified Dijkstra's algorithm to compute shortest paths and find geodesic distances concurrently, as detailed in Alg.~\ref{alg:ptu_dijkstra} (the only new lines of code to handle connections are in blue).
The final partial eigensolve was implemented using the Spectra C++ library~\cite{spectra}. If the dimension $D$ is high (i.e., larger than 100), vectorizing matrix-matrix and matrix-vector multiplications is crucial for efficiency.
\medskip

\subsection{Simulated Datasets}
We first test the performance of PTU on artificial datasets (embedded in 3D or higher) to quantifiably evaluate its behavior.\smallskip

\paragraph*{Linear Precision.} A nonlinear extension to PCA should, at the very least, be linear-precise, i.e., input data lying on a flat $d$-manifold in $\varmathbb{R}^D$ should be isometrically mapped to $\varmathbb{R}^d$. From an input pointset densely sampling a flat square shape embedded in 3D space, we compare the performance of Isomap and PTU (with $k\!=\!K\!=\!10$) in Fig.~\ref{fig:notAffinePrecise}: we visualize the error for each point based on the normalized distance between ground truth and its embedding location. As expected PTU recovers the data exactly (to numerical precision), while Isomap introduces distortions due to its use of graph-based distances.
We also use a 4D pointset that randomly samples a 3D torus in Fig.~\ref{fig:SimpleTorus}, with the embedding space being 3-dimensional. Unlike Isomap, which introduces severe distortion because of the non-convexity of the sampled domain, PTU ($k\!=\!K\!=\!10$) perfectly reproduces the torus. This last example also proves that Isomap can significantly distort data (in unexpected ways) in higher dimensions.\smallskip

\paragraph*{Non-Geodesically-Convex Domains.} An S-shaped manifold with a rectangular void (see Fig.~\ref{fig:holyS}) is well parameterized by PTU, while Isomap introduces spurious distortion associated with biased geodesic distance estimations around the void. Since this S-shaped manifold is isometrically developable, we can compare errors in parameterization on a per-point basis: Isomap reaches 15\% of relative error, while PTU (still using $k\!=\!K\!=\!10$) stays below 0.2\%. The effect of non-geodesically-convex domains is even more pronounced in higher dimensions: Fig.~\ref{fig:Petals} shows a 3D pointset sampling four petals from a surface of a sphere in 3D; this pointset is then lifted to $D\!=\!100$ dimensions and rotated by a random orthogonal transformation. The resulting non-developable and non-convex $100$D dataset is then embedded in 2D by Isomap, clearly demonstrating that graph-based distances bias results dramatically. Our algorithm, run with $k\!=\!K\!=\!10$, correctly unfurls the petal-like set.\smallskip

\begin{figure}[t]
	\includegraphics[width=\columnwidth]{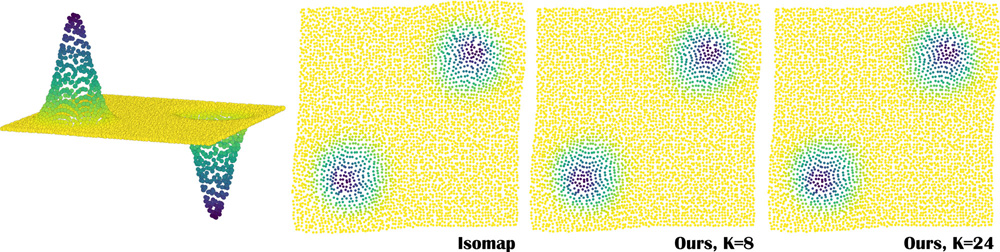}\vspace*{-2mm}
	\caption{\textbf{Gaussian landscape.} For a noise-free dense sampling of a non-developable height field (left), Isomap and PTU return comparable quasi-isometric 2D parameterizations. Using $K\!=\!8$ or $K\!=\!24$ neighbors (right) does not visually affect the result of the PTU embedding. \vspace*{-4mm}}
	\label{fig:Hills}
\end{figure}

\paragraph*{Non-developable manifolds.}  We also demonstrate our results on highly non-developable manifolds. Fig.~\ref{fig:Hills} shows a dense, noise-free pointset that is sampling a 2D height field with two Gaussian bumps. The differences between Isomap and PTU ($k\!=\!8$) are small with such a dense dataset. This example also demonstrates that the size of the geodesic neighborhood used for tangent space estimation does not dramatically affect the quality of results: PTU outputs for $K\!=\!8$ and $K\!=\!24$ are visually indistinguishable. We also test an input pointset sampling a slightly curved 3D torus in 4D: from the 3D torus in Fig.~\ref{fig:SimpleTorus}, each point $(x,y,z)$ is mapped in 4D to $(x,y,z, (x^2 + y^2)/2)$. While Isomap gets even more distorted, PTU ($k\!=\!K\!=\!10$) still produces a toroidal 3D embedding, see Fig.~\ref{fig:CurvedTorus}.\smallskip

\begin{figure}[t]
	\includegraphics[width=\columnwidth]{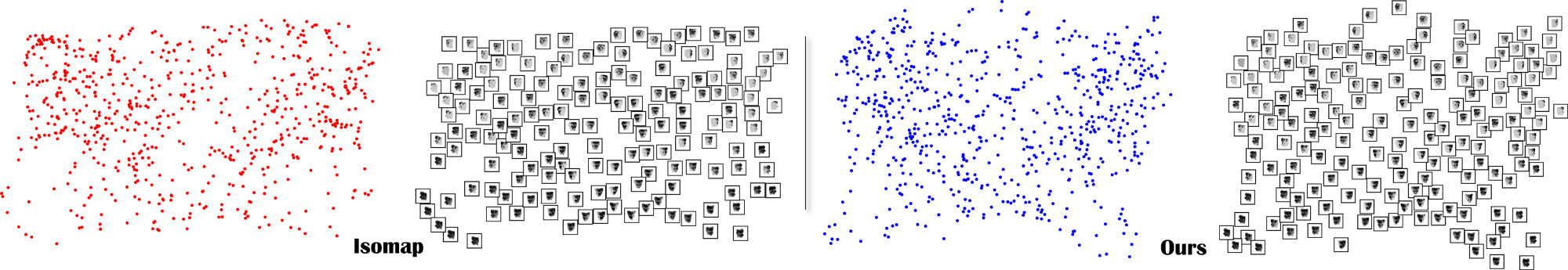}\vspace*{-2mm}
	\caption{\textbf{Faces Dataset.} For the classical Faces dataset (known to be, by construction, of intrinsic dimension $3$), the first two coordinates of both Isomap and PTU \changes{are} quite similar. Left: 2D parameterization; right: a few of the face images \changes{are shown at their actual positions} to better understand the parameterization. \vspace*{-3.5mm}}
	\label{fig:Faces}
\end{figure}

\paragraph*{Geodesic distance estimation.} We also compare the accuracy of Dijkstra, PTU, and SAKE-corrected~\cite{Max:2017} estimations of geodesic distances, by applying these 3 methods to a low-density, regular sampling of a spherical cap in Fig.~\ref{fig:improvedGeodesics} so that geodesic distances are known analytically. PTU geodesics are over 120 times more accurate than Dijkstra's, and more than 20 times better than the local geodesic correction method used in SAKE~\cite{Max:2017} (where the whole cap is treated as a single neighborhood).\smallskip

\begin{figure}[t]
  \centering
	\subfloat[Sparse noise]{\label{fig:SaltySwiss}
	\includegraphics[width=0.495\linewidth]{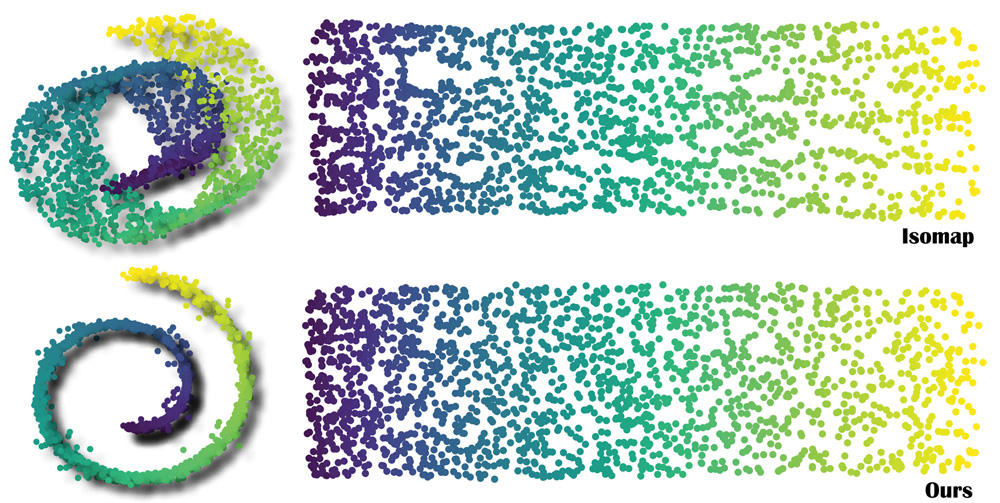}}
  \vrule 
	\subfloat[Gaussian noise]{\label{fig:NoisySwiss}\includegraphics[width=0.495\linewidth]{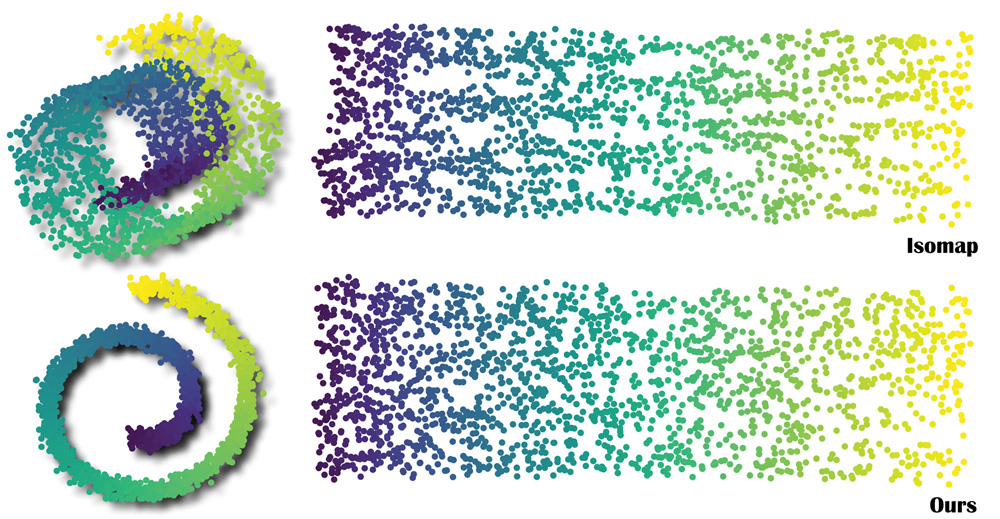}}\vspace*{-2.5mm}
	\caption{\textbf{Noisy Swiss rolls.} (a) If one adds noise to the Swiss Roll dataset in the normal direction by displacing $10\%$ of points with a uniform distribution of amplitude equal to $8\%$ of the max bounding box size and adding a Gaussian noise to the other points with standard deviation equal to $0.4\%$ of the bounding box (see two views of the 3D dataset on the left), Isomap accentuates a few low sampled regions compared to PTU.  (b) For a strong Gaussian noise (standard deviation equal to $2\%$ of the max bounding box size), Isomap suffers from clear visual artifacts while PTU returns a good parameterization without the need for robust estimations.\vspace*{-3mm}}
\label{fig:Swiss}
\end{figure}

\begin{wrapfigure}[11]{r}{0.48\textwidth}\vspace*{-7mm}
	\includegraphics[width=\linewidth]{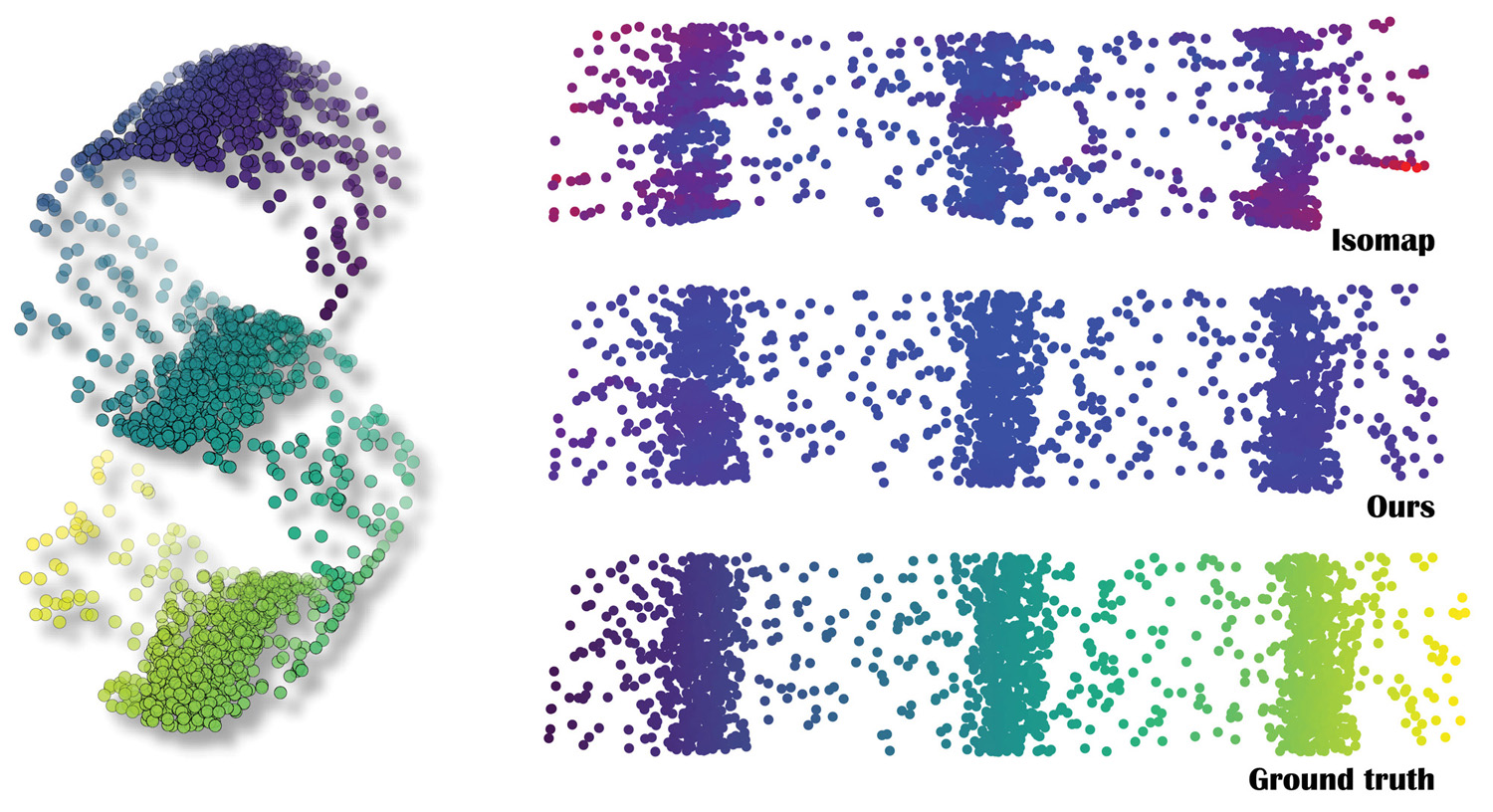}\vspace*{-3.3mm}
	\caption{\textbf{Varying Density S.}
		For a widely varying density of points (left), Isomap (top right) introduces large spurious distortions, unlike PTU (middle right).}
	\label{fig:irregularS}
\end{wrapfigure}

\paragraph*{Sensitivity to Irregular Sampling.} For a highly irregular sampling of a simple S-shaped dataset, the robustness of PTU is shown in Fig.~\ref{fig:irregularS}. Observe that Isomap completely collapses a section of the data in the right bottom corner, while PTU correctly captures the features of the input data even for a non-adapted choice of neighborhood parameters ($k\!\!=\!\!K\!\!=\!\!10$), introducing only small distortion throughout the domain. These two behaviors are visualized by the error plots using a linear color ramp from blue (0\% error) to red (10\% error).

\begin{wrapfigure}[9]{r}{0.5\textwidth}\vspace*{-7mm}\hspace{-1mm}
\includegraphics[width=\linewidth]{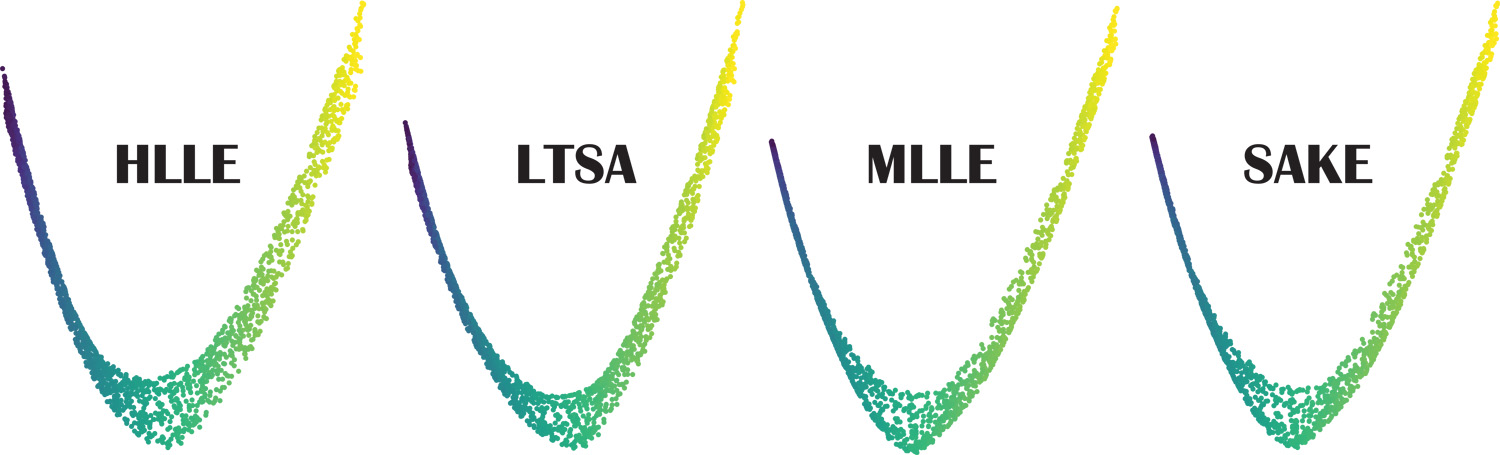}\vspace*{-3mm}
\caption{\textbf{Failure of Local Methods.} For the very noisy Swiss Roll example of Fig.~\protect\ref{fig:NoisySwiss}, none of the local manifold learning methods returns a decent parameterization, as no large intrinsic distances are exploited.}
\label{fig:localFail}
\end{wrapfigure}
\paragraph*{Sensitivity to Noise.} Figs.~\ref{fig:SaltySwiss}\,\&\ref{fig:NoisySwiss} verify that PTU (with $k\!=\!10$) is as resilient to strong noise and outliers as Isomap even without locally adapting neighborhood sizes ($K\!=\!25$ was used in the Gaussian noise case, and $K\!=\!10$ in the case of sparse noise). Local methods such as LLE~\cite{Roweis:2000:LLE},
Hessian LLE~\cite{Donoho:2003}, LTSA~\cite{LTSA}, MLLE~\cite{Zhang:2006:MLLE} or SAKE~\cite{Max:2017} (with $k\!=\!10$ for fairness of comparison) fail to unwrap noisy datasets (see Fig.~\ref{fig:localFail}) as they do not exploit the geodesic distance estimates between pairs of points that are far apart. Fig.~\ref{fig:Petals2} shows that adding noise to the petals dataset from Fig.~\ref{fig:Petals} does not alter the result of our approach significantly (we used $k\!=\!10, K\!=\!30$); yet, Isomap remain unable to unfurl the data properly, and on this seemingly simple data, all local methods fail, at times spectacularly. Please refer to the Supplemental Material for more systematic testing of the effect of noise levels on both local and global methods; as expected, PTU is systematically as good or better than all other methods.\smallskip

\paragraph{Timing.}
A performance analysis of our algorithm shows perfect agreement with the expected time complexity orders: the eigensolve dominates the computational time and scales as $O(n^{3})$, parallel transport Dijkstra scales as $O(n^{2} \log n)$, and graph construction as $O(n \log n)$, while tangent estimates are linear in $n$.  Examples of timings on an Intel i7 2GHz, 8 GB RAM laptop are 6.4s for the petals parametrization in Fig.\ref{fig:Petals},
10.9s for the irregular sampling of S-shaped manifold in Fig.~\ref{fig:irregularS}, and 9.8s for the noisy Swiss roll in Fig.~\ref{fig:NoisySwiss} (with $n=2000$). Note that these timings are for the full-blown MDS procedure, with roughly half of time spent on the other steps. Using the L-PTU variant with 0.05\%-0.1\% of the points as landmarks improved the efficiency of the MDS step by 500 to 5000 times on tested datasets, with virtually no visual difference compared to the full treatment, see Fig.~\ref{fig:L-PTU} and App.~\ref{app:SuppMat}.

\subsection{High-dimensional datasets}
We also present results on a number of real and/or high-dimensional datasets. While no ground truth is available for these examples, they allow us to compare PTU and Isomap on inputs that may not even satisfy the (single-chart) manifold assumption.\smallskip

\paragraph*{Faces Dataset.} On this classic set of 698 images of 64x64 pixels, PTU and Isomap recover the same two characteristic features of the data: Fig.~\ref{fig:Faces} shows that both arrange the images based on the azimuth and elevation of the camera, with a fairly similar global structure (2D visualization, $d\!=\!3$, $k\!=\!6$, and $K\!=\!18$).\smallskip

\paragraph*{Digit zero.} When applied to 3000 digit zero images (28x28 pixels) from the MNIST dataset, both PTU and Isomap create a parameterization of the different ways people write a zero, separating left-leaning from right-leaning and circular from oval zeros as shown in Fig.~\ref{fig:Digit0} (2D visualization, $d\!=\!4$, $k\!=\!K\!=\!10$).\smallskip

\begin{algorithm}[b]\vspace*{-0mm}
	\caption{Dimensionality reduction via parallel unfolding}\label{alg:ptu}
	\begin{algorithmic}[1]
		\Require  Pointset $\mathcal{S} = \{\mathbf{x}_i\in\varmathbb{R}^D\}_{i=1..n}$
		\State{Construct proximity graph $G$ of $\mathcal{S}$, and pairwise Dijkstra's shortest paths (Sec.~\protect\ref{sec:graph})}
		\State{Compute tangent frames $\{\mathbb{T}_i\}_{i=1..n}$ (Sec.~\protect\ref{sec:tangents})}
		\State{Compute geodesic distances $\mathbf{D}$ via parallel transport (Sec.~\protect\ref{sec:ptu_dist})}
		\State{Perform MDS on $\mathbf{D}$ to obtain $\mathbf{Z}$ (Sec.~\protect\ref{sec:computations})}
		\Ensure{Low-dimensional embedding $\mathbf{Z} =\{\mathbf{z}_i\in\varmathbb{R}^d\}_{i=1..n}$}
	\end{algorithmic}
\end{algorithm}

\begin{algorithm}[b]\vspace*{0mm}
\caption{Parallel Transport Dijkstra}\label{alg:ptu_dijkstra}
\begin{algorithmic}[1]
\Require Pointset $\mathcal{S}$, proximity graph $G$, tangent frames $\{\mathbb{T}_i\}_{i=1..n}$
\State Create minimum priority queue $P$
\For{$i \! \in \! [1, n]$}
\color{blue}
	\State $\mathbf{R}[i]\gets$ identity matrix, $\mathbf{v}[i]\gets 0$
\color{black}
	\ForAll{$\mathbf{x}_j$ adjacent to $\mathbf{x}_i$ in $G$}
		\State $Pred[j] \gets i$
		\State $dist[j] \gets |\textbf{x}_j - \mathbf{x}_i|$
		\State $P.push(dist[j], j)$
	\EndFor
	
	\While{not $P$.isEmpty()}
		\State $\mathbf{x}_{r} \gets P.\operatorname{pop\_min()}$
\color{blue}
		\State ${q} \gets Pred[r]$
		\State $\mathbf{U}S \mathbf{V}^t \gets$ SVD($\mathbb{T}_q\ \!\!^t \mathbb{T}_r$)
		\State $\mathbf{R}[r] = \mathbf{R}[q] \cdot \mathbf{U} \mathbf{V}^t$
		\State $\mathbf{v}[r] = \mathbf{v}[q] + \mathbf{R}[q]\cdot \mathbb{T}_{q}^t (\mathbf{x}_{r}  - \mathbf{x}_{q})\;\;$ \textcolor{black}{(Eq. \eqref{eq::geodFormula})}
		\State $geo\_dist[r] = |\mathbf{v}[r]|$
\color{black}		
		\ForAll{$\mathbf{x}_j$ adjacent to $\textbf{x}_r$}
			\State $temp\_dist \gets dist[r] + |\mathbf{x}_j - \mathbf{x}_r|$
			\If{$temp\_dist < dist[j]$}
				\State $dist[j] = temp\_dist$
				\State $Pred[j] = r$
				\State $P.update(dist[j], j)$ \Comment{$update()$ inserts $j\notin P$}
			\EndIf
		\EndFor
	\EndWhile
	\State $\mathbf{D}[i,1:n] \gets geo\_dist[1:n]$
\EndFor
\State Symmetrize distance matrix $\mathbf{D} \gets (\mathbf{D} + \mathbf{D}^t)/2$
\Ensure{Pairwise geodesic distance matrix $\mathbf{D} \in \varmathbb{R}^{n\times n}$}
\end{algorithmic}
\end{algorithm}

\begin{figure}[t]
	\centering \includegraphics[width=0.94\columnwidth]{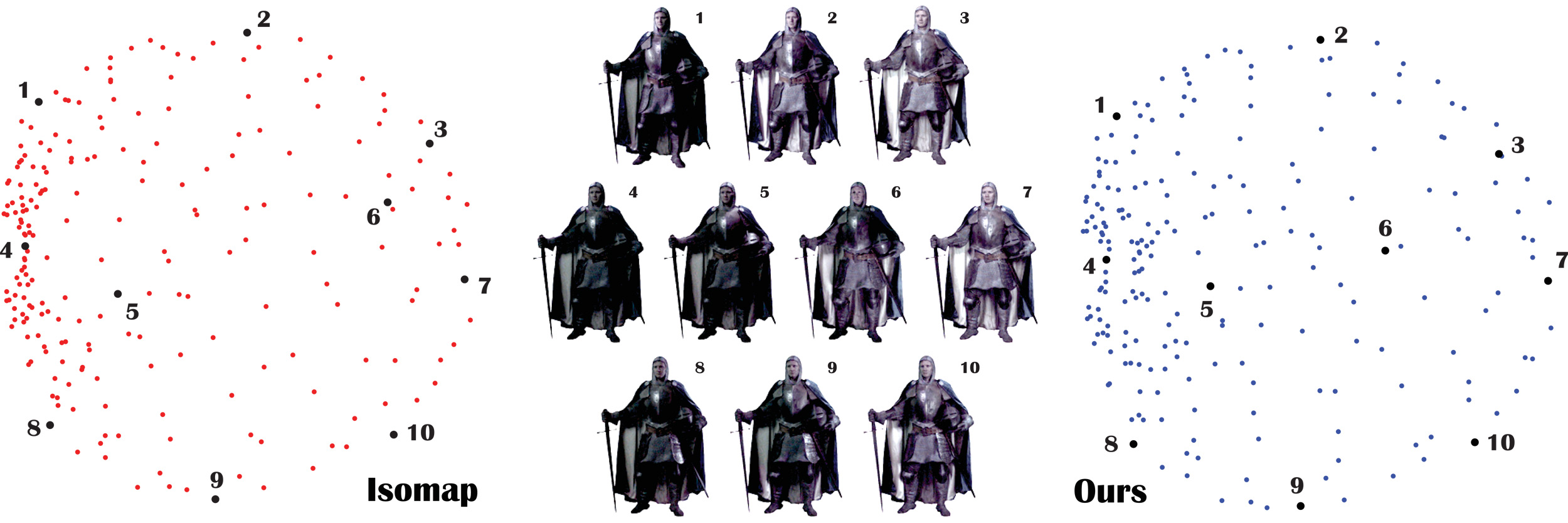}\vspace*{-3.5mm}
	\caption{\textbf{Knights.}
		From $221$ RGB images of $608 \!\times\! 456$ pixels (10 examples shown on the right) of an actor captured in full costume from different lighting directions, a 2D embedding is computed solely based on local pixel differences using Isomap and PTU (left). Both methods find a parameterization of the images corresponding to lighting direction and intensity (the knight images correspond to black dots).\vspace*{-3.5mm}}
	\label{fig:Knights}
\end{figure}

\paragraph*{Knights.} We tried our approach on the dataset from~\cite{Max:2017} of reflectance fields captured using the Light Stage apparatus~\cite{LightScape}. A static character (in a knight costume) was captured in $608 \!\times\! 456$ RGB images under 221 individual lighting directions covering a large sphere of illumination. From this pointset in $\varmathbb{R}^{831744}$, Isomap and PTU learn a flat 2D manifold that best fits this high-dimensional dataset. The result for $d\!=\!2$ and $k\!=\!K\!=\!8$, shown in Fig.~\ref{fig:Knights}, recovers positions related to light angles without any knowledge of the setup; the highly distorted results of local approaches on this dataset can be found in~\cite{Max:2017}. The black background of each image was removed for clarity.\smallskip

\paragraph*{Letter A.} From an input set of 888 images with 120x120 pixels of a rotated and resized letter `A' , the structure of this intrinsically 2-dimensional manifold in 14400 dimensional space revealed in PTU ($k\!=\!8$, $K\!=\!32$) and Isomap embeddings in Fig.~\ref{fig:letterA} (left) is very intuitive. However, when a section of the data is removed (right), Isomap suffers from its characteristic global distortion, caused by the presence of the hole; instead, the structure of PTU embedding remains basically the same.

\section{Conclusions}
The use of parallel transport on high-dimensional datasets remedies an important limitation of the Isomap approach: unfolding paths between pairs of points based on the Levi-Civita connection significantly improves the estimation of geodesic distances and removes the restriction for geodesic convexity of the input data. We demonstrated on a series of examples that our approach does indeed recover similar unfolding to Isomap for geodesically convex inputs of low- and high-dimensional data, but neither overestimates geodesic distances if large voids are present, nor suffers from large deformation in the case of non-geodesically convex inputs. This property is particularly crucial to the success of our landmark-based variant, L-PTU, which can efficiently approximates the low-dimensional embeddding of a large datasets in $\mathcal{O}(n^2 \log n)$, even in the presence of noise. \smallskip

Our work offers multiple opportunities for future research. First, connections and parallel transport are rarely used in data analysis (a notable exception being in unsupervised domain adaptation~\cite{Shrivastava:2014}), so other applications of this common concept of geometry processing may be found valuable as well, including the use of Levi-Civita connections for arbitrary, non-Euclidean metrics. Additionally, we have assumed that the input data lie on a manifold of a given dimension $d$; but real data is sometimes better captured by a CW complex, i.e., a set made out of regions of different dimensionality glued together. Being able to handle geodesic distances in this case would be a nice and useful extension. Exploiting current algorithmic work in approximating shortest paths on graphs could also be valuable to replace Dijkstra's algorithm; maybe the geodesics in heat approach~\cite{crane2013geodesics} could, similarly, be extended to high-dimensional datasets for the same purpose.
Finally, manifold learning concepts can also be extended to achieve clustering and classification of unlabeled data; using our improved geodesic distances (and potentially, its extension to CW complexes) may help provide better algorithms for these tasks, pointing to the relevance of connections in other fields of applications.

\section*{Acknowledgments}
We gratefully acknowledge the help of Mark Gillespie and Prof. Venkat Chandrasekaran in providing early feedback on our work.

\appendix
\section{Formal Statement of Proposition \changes{3}}
\label{app:proof}
Before formulating a precise error bound on our geodesic approximation to quantify Proposition \changes{3}, we first review existing results and state a few reasonable assumptions on the sampling $\mathcal{S}$ of the manifold $\mathcal{M}$. \smallskip

A similar notion of discrete parallel transport was considered by the authors of \cite{Singer:2012} in order to define and prove convergence of a connection Laplacian operator. We will make use of one of the theorems they proved towards their goal, with a proof (omitted here) relying on geometric properties of parallel transport and probabilistic guarantees (obtained through Bernstein's inequality) on the quality of tangent bases approximation. Following their notations, let $\varphi\!:\mathcal{M} \!\hookrightarrow\! \varmathbb{R}^D$ be the embedding of a smooth and compact Riemannian $d$-manifold $\mathcal{M}$ with its metric induced from the embedding space $\varmathbb{R}^D$. Points from $\mathcal{S}$ are considered to be sampled from $\mathcal{M}$ according to a probability density function $p \!\in\! \mathcal{C}^3(\mathcal{M})$. Denote the tangent bundle of $\mathcal{M}$ by $T\mathcal{M}$, the tangent space at point $\mathbf{x}_i$ by $T_i\mathcal{M}$, the differential of the embedding at $\mathbf{x}_i$ by $d\varphi_i\!:T_i\mathcal{M} \!\mapsto\! \varmathbb{R}^D$ and the parallel transport operator from $\mathbf{x}_j$ to $\mathbf{x}_j$ along a geodesic connecting them by $\mathbf{P}_{i,j}\!: T_{\!j}\mathcal{M} \!\mapsto\! T_i\mathcal{M}$. Let $\varepsilon$ be the maximum radius of the geodesic $K$-neighborhoods used to construct our approximate tangent bases $\{\mathbb{T}_i\}_{i=1}^n$ (see Eq. \eqref{eq:defFrames} and its description) and finally, let $\varepsilon_g>0$.\\
\noindent\emph{\textbf{Theorem} (from~\cite{Singer:2012}).
Assume $\varepsilon = \mathcal{O}(n^{-\frac{1}{d+2}})$ and consider two points $\mathbf{x}_i$ and $\mathbf{x}_j$ of $\mathcal{M}$, such that the geodesic distance between them is $\mathcal{O}(\varepsilon_g)$. Then for any vector $\mathbf{u}_j \in T_j\mathcal{M}$, with high probability:\vspace*{-2mm}
\begin{equation}
\mathbf{R}_{i,j} \mathbb{T}_j^t d\varphi_j \left[\mathbf{u}_j\right] = \mathbb{T}_i^t d\varphi_i \left[\mathbf{P}_{i,j} \mathbf{u}_j\right] + \mathcal{O}(\gamma_{i,j})
\label{eq:transportConv}\vspace*{-2mm}
\end{equation}
where $\gamma_{i,j} \!=\! \varepsilon^{3} \!+\! \varepsilon_g^{3}$ if $\mathbf{x}_i$ and $\mathbf{x}_j$ are at least $\varepsilon$-away from the boundary of $\mathcal{M}$ and $\gamma_{i,j} \!=\! \varepsilon \!+\! \varepsilon_g^{3}$ otherwise.}\smallskip

\noindent In addition to this theorem, we will use two more assumptions:
\begin{enumerate}
	\item[I.] Let $\mathbf{c}_p \!=\! (\mathbf{x}_{i_1},..., \mathbf{x}_{i_m})$ be the shortest polyline connecting $\mathbf{x}_{i_1}$ and $\mathbf{x}_{i_m}$ in the proximity graph $G$. First, we assume that the polyline is included in an $\varepsilon_d$-thickening of the real geodesic $\mathbf{c}_g$ connecting the same endpoints, where $\varepsilon_d$ is a positive constant such that $\varepsilon_d^2\kappa_s \ll 1$, with $\kappa_s$ denoting the maximum absolute value of intrinsic sectional curvature of $\mathcal{M}$. Note that this condition is less stringent than the one used to prove convergence of Dijkstra polylines to geodesic curves~\cite{isomap}: sampling voids of maximum geodesic diameter $\varepsilon_d$ are allowed in the input pointset $\mathcal{S}$. 
	
	\item[II.] Let $\mathbf{g}_{i_s}$ be a tangent vector in $T_{\!i_s}\mathcal{M}$ that connects the endpoints of the Cartan development of the geodesic curve between $\mathbf{x}_{i_{s}}$ and $\mathbf{x}_{i_{s+1}}$ onto $T_{\!i_s}\mathcal{M}$. Denoting $\mathbf{e}_{i_s} \!=\! (\mathbf{x}_{i_{s+1}\!} \!-\! \mathbf{x}_{i_s})$, we will also assume:\vspace*{-2mm}
	\begin{equation}
	\mathbb{T}_{i_s}^t d\varphi_{i_s} \mathbf{g}_{i_s} = \mathbb{T}_{i_s}^t \mathbf{e}_{i_s} + \mathcal{O}(\gamma_{i_s,i_{s+1}})\vspace*{-2mm}
	\label{eq::condition}
	\end{equation}
	This condition links local curvature and sampling density: the projection of $\mathbf{e}_{i_s}$ onto the approximate tangent basis $\mathbb{T}_{i_s}$ must be close to the unwrapped geodesic in the same basis. Note that the length rescaling step (Eq.~\eqref{eq::lengthRescaling}) can help in practice to tighten the error bound $\mathcal{O}(\gamma_{i_s,i_{s+1}})$.
\end{enumerate}\smallskip

\noindent\emph{\textbf{Proposition \changes{3} revisited:} Under assumptions I, II and the assumptions of the theorem, the PTU estimate $d_\text{\tiny PTU}$ of the geodesic distance between points $\mathbf{x}_{i_1}$ and $\mathbf{x}_{i_m}$, based on a Dijkstra shortest polyline $\mathbf{c}_p \!=\! (\mathbf{x}_{i_1},..., \mathbf{x}_{i_m})$ on $G$, provides an approximation of the length $d_{g}$ of the geodesic curve $\mathbf{c}_g$ with the same endpoints in the following sense:\vspace*{-2mm}
\[
d_\text{\tiny PTU}(\mathbf{x}_{i_1}, \mathbf{x}_{i_m}) = d_{g}(\mathbf{x}_{i_1}, \mathbf{x}_{i_m}) + \mathcal{O}(\delta),\vspace*{-2mm}
\]
where $\mathcal{O}(\delta)$ is between $\mathcal{O}(m^2(\varepsilon +\varepsilon_g^3+\varepsilon_g \varepsilon_d^2\kappa_s) )$ and $\mathcal{O}(m^2(\varepsilon^3 + \varepsilon_g^3+\varepsilon_g \varepsilon_d^2\kappa_s))$ depending on how many polyline segments are close to the manifold boundary.}\smallskip

\emph{Proof:} First, observe that condition \eqref{eq::condition} implies:\vspace*{-2mm}
\[
\mathbf{v}_{i_s}= \left(\prod_{j=1}^{r-1} \mathbf{R}_{i_j,i_{j+1}} \right) \biggl[\mathbb{T}^{\ t}_{i_{r}} \mathbf{e}_{i_s} \biggr] = \left(\prod_{j=1}^{r-1} \mathbf{R}_{i_j,i_{j+1}} \right) \biggl[\mathbb{T}^{\ t}_{i_{r}} \mathbf{g}_{i_s} \biggr] + O(\varepsilon_g^3)\vspace*{-2mm}
\]
Using Eq.~\eqref{eq:transportConv} repeatedly, we obtain that $\mathbf{v}_{i_s}$ is approximately equal to $\mathbf{g}_{i_s}$ parallel-transported to $\mathbb{T}_{i_1}$ along a piecewise-geodesic curve $\mathbf{c}_{pg}$ passing through the points $(\mathbf{x}_{i_1},..., \mathbf{x}_{i_m})$, i.e.:\vspace*{-2mm}
\[
\mathbf{v}_{i_s} = \mathbb{T}_{i_1} d\varphi_{i_1} \left(\prod_{j=1}^{s-1} \mathbf{P}_{i_j,i_{j+1}} \right) \biggl[ \mathbf{g}_{i_s}\biggr] + O\left(\sum_{j=1}^{s-1} \gamma_{i_j,i_{j+1}}\right) + O(\varepsilon_g^3)\vspace*{-2mm}
\]
Denoting the first term of the right-hand side by $\mathbf{w}_{i_s}$ and calling $\mathbf{w} = \sum_{s=1}^m \mathbf{w}_{i_s}$, we obtain:\vspace*{-2mm}
\[
\sum_{s=1}^m \mathbf{v}_{i_s} = \sum_{s=1}^m \mathbf{w}_{i_s} + O(\gamma)\text{ , i.e.,} \quad \mathbf{v} = \mathbf{w} + \mathcal{O}(\gamma)\vspace*{-2mm}
\]
\begin{wrapfigure}[13]{r}{0.35\textwidth}\vspace*{-5.5mm}\hspace*{-3mm}
  \centering
  \includegraphics[width=1.02\linewidth]{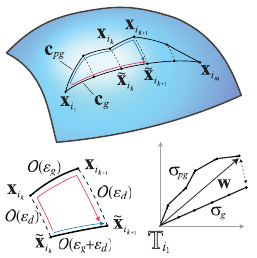}
\end{wrapfigure}
where $\mathcal{O}(\gamma)$ takes value between $\mathcal{O}(m^2(\varepsilon + \varepsilon_g^3))$ and $\mathcal{O}(m^2(\varepsilon^3 + \varepsilon_g^3))$ depending on how many polyline segments are close to the boundary; note that $\mathbf{w}$ is the vector connecting the endpoints of Cartan's development $\sigma_{pg} \in T_{i_1}\mathcal{M}$ of the \emph{piecewise-geodesic} curve $\mathbf{c}_{pg}$ interpolating the polyline vertices (see inset, bottom right). 
To show that the curve $\sigma_{pg}$ stays close to Cartan's development $\sigma_g$ of the geodesic $\mathbf{c}_g$, recall that for a patch of $\mathcal{M}$ with diameter $\mathcal{O}(\varepsilon_d)$, the change in direction of a vector parallel-transported along different paths within the patch is bounded by $\mathcal{O}(\kappa_s \varepsilon_d^2)$ (this result relies on a bounded \textit{curvature transformation}, which in turn follows from our bound on absolute value of \textit{sectional curvature} from Assumption I). Using Assumption II, we can cut the geodesic $\mathbf{c}_g$ connecting $\mathbf{x}_{i_1}$ and $\mathbf{x}_{i_m}$ into curved segments with vertices $\tilde{\mathbf{x}}_{i_k}$, such that $d_g(\tilde{\mathbf{x}}_{i_k},\mathbf{x}_{i_k}) \!=\! \mathcal{O}(\varepsilon_d)$ for $k\!=\!2,\hdots,m\!-\!1$ (we set $\tilde{\mathbf{x}}_{i_1} = {\mathbf{x}}_{i_1}$ and $\tilde{\mathbf{x}}_{i_m} = {\mathbf{x}}_{i_m}$ for notational consistency). Because the geodesic length of $\mathbf{x}_{i_{k+1}}\mathbf{x}_{i_k}$ is $\mathcal{O}(\varepsilon_g)$, the diameter of a patch enclosed by four geodesic segments $\mathbf{x}_{i_k} \mathbf{x}_{i_{k+1}}$, $\tilde{\mathbf{x}}_{i_{k+1}}\tilde{\mathbf{x}}_{i_{k}}$, $\tilde{\mathbf{x}}_{i_k} \mathbf{x}_{i_k}$ and $\mathbf{x}_{i_{k+1}}\tilde{\mathbf{x}}_{i_{k+1}}$ is $\mathcal{O}(\varepsilon_g \!\!+\!  \varepsilon_d) \!=\! \mathcal{O}(\varepsilon_d)$ assuming $\varepsilon_g \lesssim \varepsilon_d$ (see inset). Thus the result of parallel transporting a vector along 3 geodesic segments $\tilde{\mathbf{x}}_{i_k}\mathbf{x}_{i_k}$,  $\mathbf{x}_{i_k} \mathbf{x}_{i_{k+1}}$ and $\mathbf{x}_{i_k}\tilde{\mathbf{x}}_{i_k}$ deviates in direction from parallel transporting the same vector along geodesic segment $\tilde{\mathbf{x}}_{i_k}\tilde{\mathbf{x}}_{i_{k+1}}$ by $\mathcal{O}(\kappa_s\varepsilon_d^2)$. Summing up contributions from all the patches for $k=1,\hdots,\ell$, and taking into account cancellations from segments $\mathbf{x}_{i_k}\tilde{\mathbf{x}}_{i_k}$ and $\tilde{\mathbf{x}}_{i_k}\mathbf{x}_{i_k}$ for $k\!=\!2,\hdots,\ell\!-\!1$, the directional error in parallel transporting a vector along the piecewise geodesic curve $\mathbf{x}_{i_1}\mathbf{x}_{i_2}\hdots\mathbf{x}_{i_\ell}\tilde{\mathbf{x}}_{i_\ell}$ compared to its transport along the true geodesic segment $\mathbf{x}_{i_1}\tilde{\mathbf{x}}_{i_\ell}$ becomes $\mathcal{O}(\ell \kappa_s \varepsilon_d^2)$. As a result, performing Cartan's development patch by patch, the final discrepancy between the endpoint positions of $\sigma_{pg}$ and $\sigma_g$ incurred by developing them using parallel transport along piecewise-geodesic curve $\mathbf{c}_{pg}$ vs. the true geodesic $\mathbf{c}_g$ is $\mathcal{O}(m^2 \kappa_s \varepsilon_g \varepsilon_d^2)$, as it combines the cumulative directional errors of parallel transported tangent vectors and the lengths of the corresponding segments. 
\smallskip

Given that Cartan's development of a geodesic curve is a straight segment, we conclude that $||\mathbf{w}||_2 = d_{g}(\mathbf{x}_{i_1}, \mathbf{x}_{i_m}) + \mathcal{O}(m^2 \kappa_s \varepsilon_g \varepsilon_d^2)$. By construction, we have $d_\text{\tiny PTU}(\mathbf{x}_{i_1}, \mathbf{x}_{i_m}) = ||\mathbf{v}||_2$, implying that:\vspace*{-1mm}
\[
d_\text{\tiny PTU}(\mathbf{x}_{i_1}, \mathbf{x}_{i_m}) = d_{g}(\mathbf{x}_{i_1}, \mathbf{x}_{i_m}) + \mathcal{O}(\delta),\vspace*{-1mm}
\]
where $\mathcal{O}(\delta) = \mathcal{O}(m^2 \varepsilon_g \varepsilon_d^2\kappa_s \!+\! \gamma)$.  Note that our discrete unfolding of \emph{any} polyline converges to the corresponding Cartan's development; however, in general it has to be a (nearly-)shortest polyline for its extremities to be at a distance approximating the proper geodesic length, as geodesic curves are only locally (and not globally) shortest.
\hfill\ensuremath\blacksquare\\[-3mm]

Finally, we note that while our assumptions are weaker that the ones used to prove convergence of Dijkstra polylines to geodesic curves~\cite{isomap}, the error for graph-based distance approximations has linear dependence on the number of polyline segments, while our bound depends quadratically on the number of segments. However, in practice our parallel transport based method consistently outperforms Dijkstra path approximation (see Fig.~\ref{fig:improvedGeodesics}), potentially pointing to the existence of tighter bounds.

\section{Additional Results}
\label{app:SuppMat}

This last appendix contains additional results to provide further tests of the Parallel Transport Unfolding (PTU) approach and comparisons to other existing Manifold Learning algorithms.

\subsection{PTU vs. Local Methods}
One of the key advantages of Parallel Transport Unfolding is its resilience to noise: like Isomap~\cite{isomap}, PTU uses \emph{all} geodesic distances to embed a dataset into a low-dimensional space, which allows for a much increased robustness to noise and outliers. In this section, we provide further numerical tests to confirm this statement.

\begin{figure}[h]\vspace*{-1mm}\centering
\includegraphics[width=0.9\linewidth]{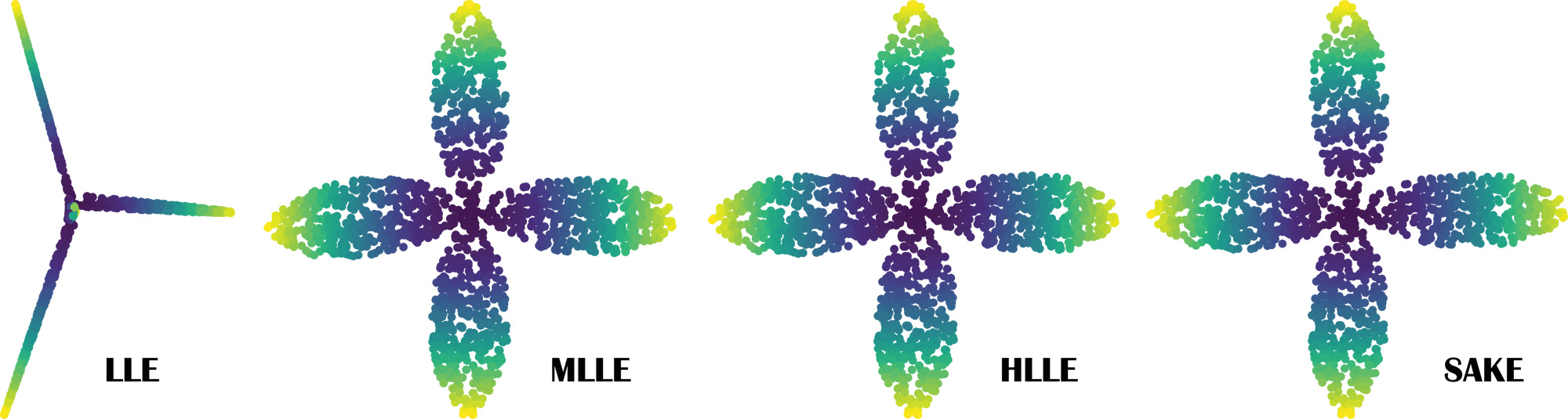}\vspace*{-3mm}
	\caption{\textbf{Local Methods for Noiseless Petals.} From a 3D sampling of 4-petal shaped portion of a sphere (see Fig.~\ref{fig:Petals}), local methods such as Modified LLE~\cite{Zhang:2006:MLLE}, Hessian LLE~\cite{Donoho:2003}, or SAKE~\cite{Max:2017} have no issue with the non-convexity of the intrinsic geometry (unlike Isomap), and give results nearly equivalent to PTU. A notable exception is LLE~\cite{Roweis:2000:LLE}, which returns a near degenerate solution. \vspace*{-2mm}}
	\label{fig:SMPetalsLocal}
\end{figure}

\paragraph{Petals dataset.}
First off, the noiseless Petals dataset makes clear that Isomap fails due to the obvious geodesic non-convexity while PTU has no problem finding the proper four petals as demonstrated in Fig.~\ref{fig:Petals}. In this noiseless case, it turns out that most local methods, such as Modified LLE~\cite{Zhang:2006:MLLE}, Hessian LLE~\cite{Donoho:2003}, and SAKE~\cite{Max:2017}, do also remarkably well (see Fig.~\ref{fig:SMPetalsLocal}), as convexity (or lack thereof) plays basically no role in their embeddings.

However, if a bit of noise is added to this example, local methods fail (at times spectacularly): for a moderate Gaussian noise with a standard variation of $3\%$ of the radius of the sphere on which the petals lie, Fig.~\ref{fig:Petals2} shows that local methods all fail while PTU keeps a very similar embedding since it relies on all pairwise geodesic distances---Isomap too to a certain extent, even if it is clearly more deformed than in the noiseless case. 

\begin{figure}[htb]
  \centering 
	\includegraphics[width=0.9\linewidth]{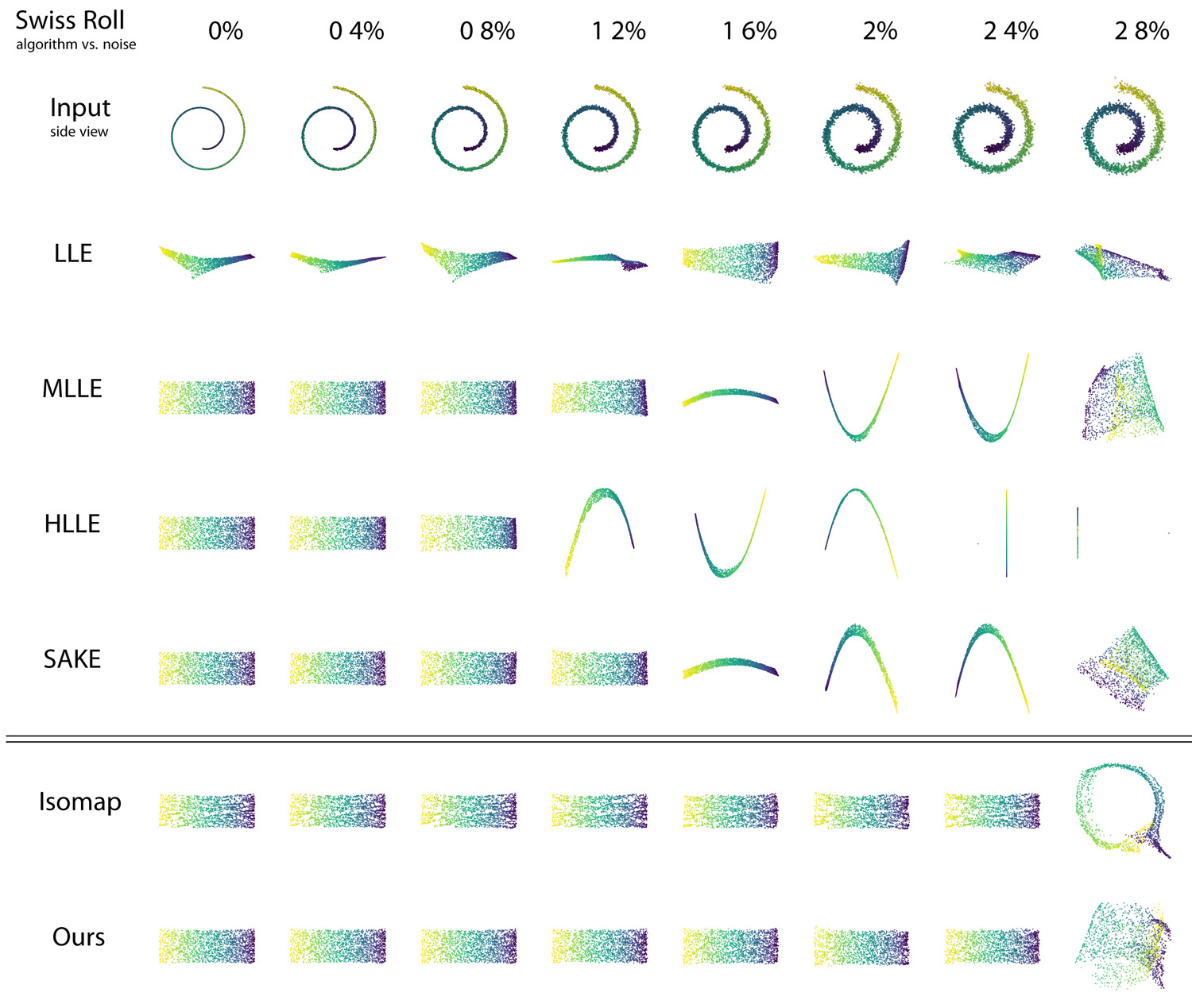}\vspace*{-3mm}
  \caption{\textbf{Effects of noise on local and global methods.} Using the Swiss Roll dataset, Gaussian noise with standard deviation given as a percentage of the bounding box of the original noiseless swiss roll is added along the normal. We use the same number of neighbors (10) for local methods to provide a fair comparison (it prevents shortcutting as much as possible; using larger values would make the local methods fail earlier). Local methods all failed around $\sigma=1.3\%$, while global methods (Isomap and our approach) fare well until $2.7\%$. At 2.8\%, the neighbors of a datapoint may belong to several different branches of the roll, which makes it impossible even for global methods to handle.  \vspace*{-3mm}}\label{tab:Noise}
\end{figure}

\paragraph{Study of noise effects on local and global methods.}
In order to better demonstrate the robustness of global methods to noise, we use the simple (and very widely used) Swiss Roll dataset, and use both local and global methods on this dataset with an increasing amount of Gaussian noise along the normal of this roll (varying the standard deviation from 0 to $2.8\%$ of the bounding box size). The number of neighbors is set to 10 for all methods to offer a fair comparison. As Fig.~\ref{tab:Noise} clearly shows, the best local method (SAKE on this example) starts failing at half of the maximum standard deviation that global methods can handle. When the standard deviation of the noise reaches $2.8\%$, even global methods fail as the proximity graph starts having numerous connections across branches: pairwise geodesic distances will have too many incorrect values to be able to recover a decent embedding. Note that at this level of noise, the dataset is far from the manifold assumption we are making about input data.

\begin{figure}[thb]
	\includegraphics[width=\linewidth]{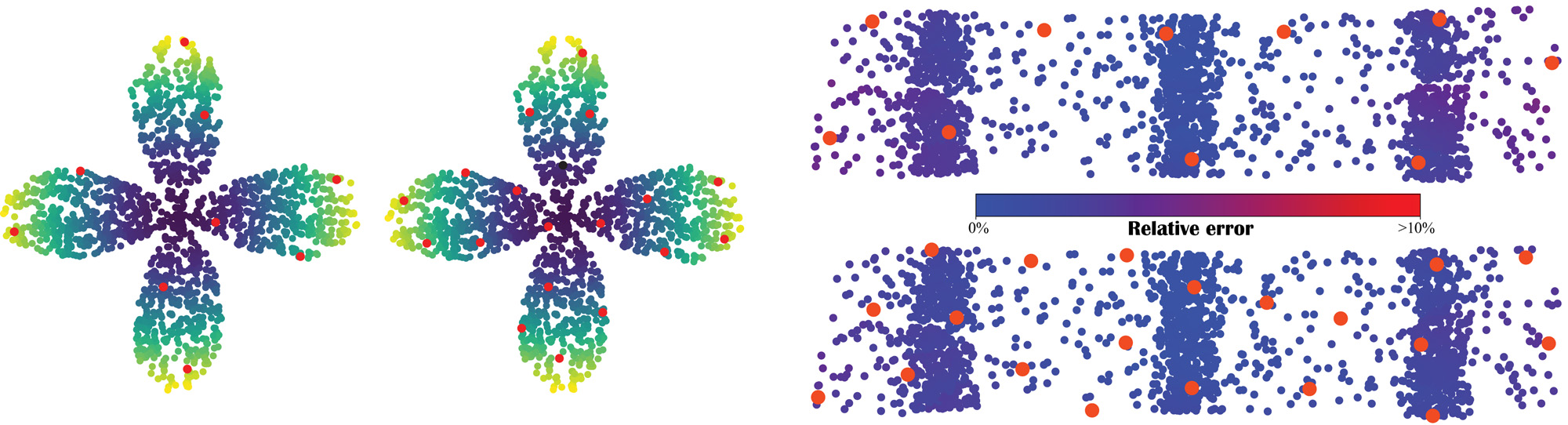}\vspace*{-3mm}
	\caption{\textbf{Landmark-PTU.} Landmarks are colored red for clarity. (Left) the use of 9 landmarks (left) or 19 landmarks (right) is enough to reconstruct the petals in the noiseless example of Fig.~\ref{fig:Petals}. (Right) using 10 landmarks (top) vs. 20 landmarks (bottom) is visually very similar on this 2000-point datasets, although one can notice a slight distortion as indicated by the color of the mapped points using the color ramp based on distortion error compared to the expected perfect embedding.\vspace*{-2mm}}
	\label{fig:SML-PTU}
\end{figure}

\subsection{Landmark-PTU}
Fig.~\ref{fig:SML-PTU} provides more results for L-PTU: in order to complement Fig.~\ref{fig:L-PTU}, we also provide the results for landmarks on the \emph{noiseless} Petals dataset, as well as on the highly-irregular S-shaped dataset from Fig.\ref{fig:irregularS}. Here again, less than $1\%$ of landmarks is enough to capture the shape almost perfectly.

{\small
\bibliographystyle{apalike}
\bibliography{GeoIsoMap}
}

\end{document}